\documentclass[11pt]{article}

\usepackage[preprint]{acl}

\usepackage{times}
\usepackage{latexsym}
\usepackage{amsmath}

\usepackage[T1]{fontenc}

\usepackage[utf8]{inputenc}

\usepackage{microtype}

\usepackage{inconsolata}

\usepackage{graphicx}
\usepackage{xspace}
\usepackage{booktabs}

\usepackage{cleveref}
\usepackage{tcolorbox}
\usepackage{verbatim}

\usepackage{makecell}
\usepackage{multirow}

\usepackage{array}
\usepackage{longtable}
\usepackage{booktabs}
\usepackage{ltablex}

\title{
  \raisebox{-0.25\height}{\includegraphics[height=1.5em]{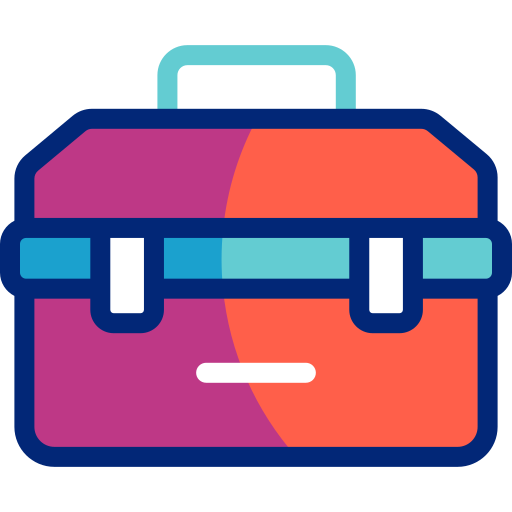}}
  \hspace{0.5em}
  \textbf{\texttt{AITutor-EvalKit}: Exploring the Capabilities of AI Tutors}
  \vspace{-1em}
}

\author{
 \textbf{Numaan Naeem}\thanks{\xspace Equal contribution.},
 \textbf{Kaushal Kumar Maurya}$^*$,
 \\
 \textbf{Kseniia Petukhova} and 
 \textbf{Ekaterina Kochmar}
\\
Mohamed bin Zayed University of Artificial Intelligence, Abu Dhabi, UAE
\\
 \small{
   \texttt{\{Numaan.Naeem, kaushal.maurya, kseniia.petukhova, ekaterina.kochmar\}@mbzuai.ac.ae}
 }
}

\begin{document}
\maketitle
\begin{abstract}
We present \texttt{AITutor-EvalKit}, an application that uses language technology to evaluate the pedagogical quality of AI tutors, provides software for demonstration and evaluation, as well as model inspection and data visualization. This tool is aimed at education stakeholders as well as *ACL community at large, as it supports learning and can also be used to collect user feedback and annotation.

\end{abstract}

\section{Introduction}
\label{sec:intro}

Personalized one-on-one tutoring has long been recognized as a highly effective educational approach~\cite{Bloom1984The2S}. Yet, its widespread adoption is constrained by the limited availability of qualified tutors \cite{wang2024tutor} and the high costs associated with tutor training~\cite{Kelly2020UsingGO}, among other impediments~\cite{Yoon2007ReviewingTE,Boyd2008TeacherPA}. An alternative to human tutoring is provided by AI tutoring systems, especially those relying on recent advances in large language models (LLMs), such as Khanmigo \cite{khan2024khanmigo} and Tutory.\footnote{\url{https://tutory.io}} Despite the remarkable success of LLMs in various tasks~\cite{minaee2024large}, their adoption in education is hindered by lack of a clear understanding of what these models are capable of~\cite{tack-etal-2023-bea,jurenka2024towards} and how pedagogically useful they are~\cite{macina-etal-2023-opportunities}, which results in lack of trust on the part of key educational stakeholders. With the fast development of LLMs and their easy integration into learning tools, questions about the evaluation of AI-driven tutor performance become increasingly relevant~\cite{kosmyna2025your}. The goal of our tool is two-fold: (1) via the open-source and open-access code, we provide a practical, customizable and versatile evaluation tool that can be applied to a variety of educational scenarios; and (2) via informative demonstrations, we aim to raise awareness of the stakeholders (e.g., students and teachers) as well as practitioners interested in the state of the art in AI in Education (i.e., *ACL community at large) of the current capabilities of LLMs-as-tutors. Although our tool is an \textit{early research prototype}, through its extendable nature, we aim to facilitate research in this exciting and emerging area of applied NLP research.

\begin{figure}[!t]
    \centering
   \includegraphics[width=0.9\linewidth]{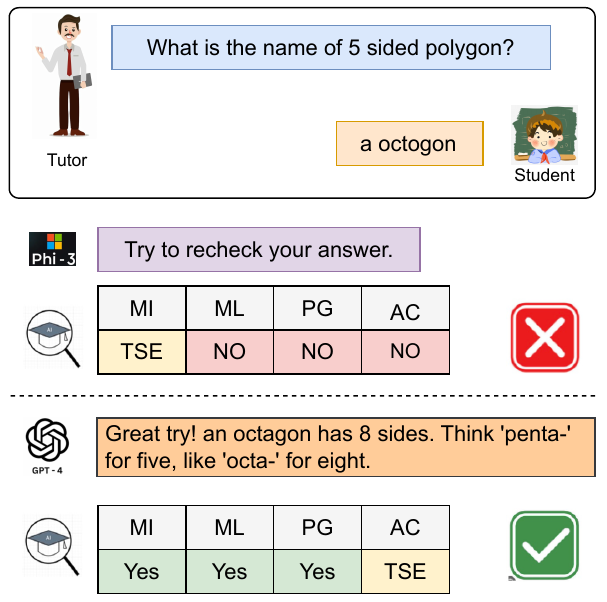}
    \caption{\small This example shows a sample dialogue and its pedagogical-ability evaluation by the \textbf{{\tt LoMTL}} model using the \texttt{AITutor-EvalKit}. The evaluation follows the four dimensions from \citet{kochmar-etal-2025-findings}: \textbf{MI} (Mistake Identification), \textbf{ML} (Mistake Location), \textbf{PG} (Providing Guidance), and \textbf{AC} (Actionability). \textbf{TSE:} To some extent.  
}\vspace{-0.4cm}
    \label{fig:example}
\end{figure}

The current version of our tool focuses on the pedagogical quality evaluation of tutor responses in the context of Student Mistake Remediation (SMR)~\cite{boaler2013ability,handa-etal-2023-mistakes} in the mathematical domain, where tutors address errors or misconceptions that hinder students' progress~\cite{wang-etal-2024-bridging, wang2024tutor, macina-etal-2023-mathdial}. As the foundation for evaluation, we use the established taxonomy from \citet{maurya-etal-2025-unifying}, which is grounded in the learning sciences principles and which allows us to assess the quality of tutor responses along four key SMR dimensions~\cite{kochmar-etal-2025-findings}: (1) {\em mistake identification}, concerned with whether tutor's response notifies the student of the committed mistake; (2) {\em mistake location}, focusing on whether the tutor clearly points to the erroneous part in the student's solution; (3) {\em providing guidance}, evaluating the quality of the pedagogical guidance; and (4) {\em actionability}, assessing whether tutor's response makes it clear what the student should do next. These dimensions are further outlined in Table~\ref{tab:detailed_def}, with an illustrative example provided in Figure~\ref{fig:example}.

The structure and implementation of the front- and backend of our tool are described in Section~\ref{sec:system}. Using {\tt MRBench} dataset~\cite{maurya-etal-2025-unifying} and taking inspiration from the BEA 2025 shared task on AI tutor response evaluation~\cite{kochmar-etal-2025-findings}, we introduce a novel, efficient, and light-weight multi-task learning model that addresses the four dimensions of pedagogical quality evaluation ($\S$\ref{sys:mtl}). The outputs of the model, as well as those of an open-source ({\tt Prometeus}~\cite{kim-etal-2024-prometheus}) and a commercial ({\tt GPT-5}) LLMs used as judges, are displayed using an interactive browser-based UI with helpful visualizations ($\S$\ref{sys:demo}). The evaluation results ($\S$\ref{sec:eval}) suggest that while our model achieves competitive results when evaluated against gold standard annotation, its outputs are also perceived by users to be at least as accurate as those of the commercial LLM-as-judge models, and the UI is considered informative and easy to use. 
We present and publicly release:
\vspace{-0.5em}
\begin{itemize}
\item The first of its kind, open-access and open-source model aimed at evaluation of the pedagogical quality of AI tutor responses available at MIT-licensed python repository: \url{https://github.com/kaushal0494/AITutor-EvalKit}. We believe it to be useful for AI-in-Education practitioners and developers, as it is highly customizable, allowing researchers to apply it to further educational contexts and dialogues, as well as to extend it to other scenarios and domains.\vspace{-0.5em}
\item An interactive UI available at \url{https://demo-ai-tutor.vercel.app}, which communicates the results and showcases the capabilities of AI tutors in an interpretable way, which we consider to be of interest to education stakeholders and the *ACL community at large. The demo tool can also be run locally with the user's own data and models following the instructions provided in the GitHub repository.\vspace{-0.5em}
\item A short video demonstrating the tool available at \url{https://www.youtube.com/watch?v=9qgDfrhzOvg}.
\end{itemize}
\vspace{-0.5em}

\section{Related Work}


Over the years, the NLP community has seen significant advances in the development of publicly available toolkits for modeling and evaluation, including Hugging Face Transformers \cite{wolf2019huggingface}, NLTK \cite{bird2006nltk}, and Scikit-learn \cite{pedregosa2011scikit}. These toolkits have enhanced code reusability, enabling researchers to focus on developing more sophisticated models and metrics. 

However, in the educational domain, especially in conversational dialogues, there remains a lack of robust tools to push research boundaries. Some progress has been made with frameworks like ConvoKit \cite{chang2020convokit}, which facilitates the manipulation and analysis of general conversational data, and the social interactions embedded within. More recently, Edu-ConvoKit \cite{wang-demszky-2024-edu} was developed for preprocessing, annotation, and analysis specifically for educational dialogues. While these toolkits contribute significantly to data handling and analysis, they fall short of addressing the evaluation of pedagogical quality in AI-driven educational systems. To the best of our knowledge, there is no toolkit that supports \textit{on-the-fly} assessment of the pedagogical abilities of AI tutors. With \texttt{AITutor-EvalKit}, we aim to fill this critical gap by providing an open-source toolkit for the systematic evaluation of AI tutors. The toolkit integrates a popular taxonomy proposed by \citet{maurya-etal-2025-unifying} into an automated evaluation framework. It is also designed to be extensible to additional evaluation aspects such as the ones proposed by \citet{macina-etal-2025-mathtutorbench}, who augment pedagogical assessment with measures of student expertise and understanding. By supporting and unifying such evaluation methods, \texttt{AITutor-EvalKit} aims to facilitate progress in this underexplored yet important research area.

\begin{figure*}[!htb]
    \centering
    \includegraphics[width=1\linewidth]{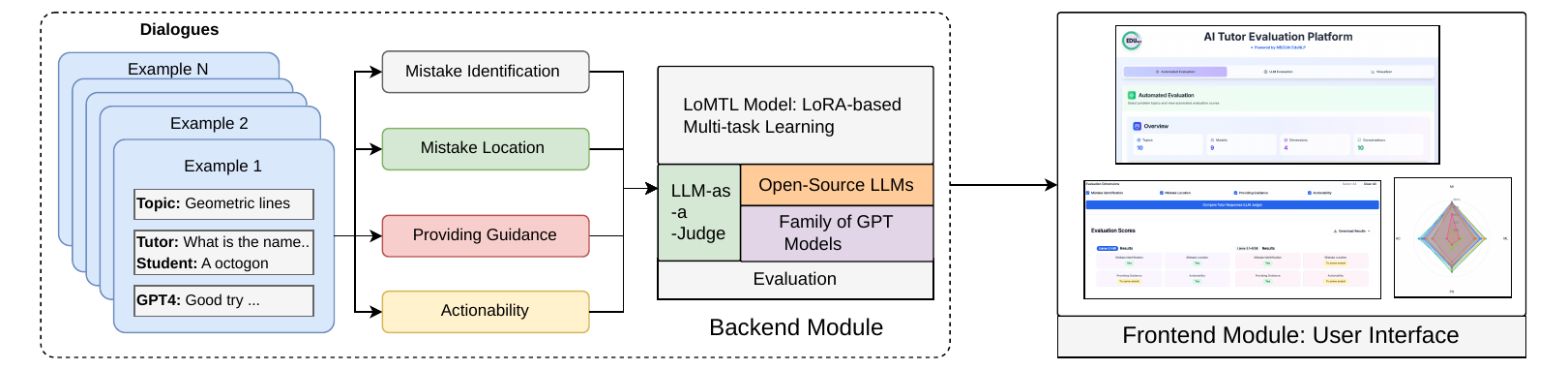}
    \caption{\texttt{AITutor-EvalKit} pipeline: The backend module includes several model options to assess the pedagogical soundness of tutors' responses, and the frontend presents evaluation outputs in an interactive user interface.}
    \label{fig:demopipe}
\end{figure*}

\section{System and Demonstration Description}
\label{sec:system}



{\tt AITutor-EvalKit} consists of two major modules: \textbf{backend} and \textbf{frontend}, with the pipeline illustrated in Figure~\ref{fig:demopipe}. The backend module includes several models for tutor response evaluation, and frontend seamlessly integrates these evaluation outputs with an interactive, customizable, and flexible UI. Different audience groups can benefit from different functionalities of the toolkit: e.g., researchers and developers can use both modules, train their own automated models, use their own datasets, and launch the demo UI locally, while teachers,  policymakers and other educational stakeholders can use the frontend module to understand the capabilities of LLMs-as-tutors and make decisions accordingly.

\subsection{Backend Module: Evaluation Models}
\label{sys:mtl}


The backend module has two components: \textit{a specialized automated evaluation model} and \textit{an evaluation pipeline for LLMs-as-judges}. In this section, we provide details on the task, the training data (for the automated evaluation model), the test data, and the automated evaluation model itself, as well as the usage of an open-source LLM and a closed-source GPT model as judges.

\subsection{Student Mistake Remediation Task}

This task considers mathematical educational dialogues between a student and a tutor, where interactions are driven by student's mistakes or confusions, and the AI tutor aims to remediate them through pedagogically appropriate responses. Formally, let the conversation history be $H = \{(T_1, S_1), (T_2, S_2), \dots, (T_t, S_t)\}$, where $T_i$ and $S_i$ denote the tutor's and student's $i$-th utterances, respectively. Let $S_k$, $k \in [1, \dots, t]$, denote the most recent student's utterance containing a mistake or confusion; the tutor then produces $T_{t+1}$ to address it. The proposed toolkit evaluates the pedagogical quality of $T_{t+1}$ along eight dimensions defined by \citet{maurya-etal-2025-unifying}.

\subsubsection{Dataset}

As discussed in Section~\ref{sec:intro}, we used  {\tt MRBench} dataset by \citet{maurya-etal-2025-unifying}, which has 491 dialogues (300 in the development set and 191 in the test set), each paired with seven LLM-generated tutor responses and one or two human tutor responses. Each response is annotated by human annotators for each of the four dimensions using the categories ``Yes,'' ``To some extent,'' and ``No.'' We also created a randomly selected \textit{demonstration set} consisting of 10 dialogues, which is a subset of the test set. More details on the dimensions, annotation, and data statistics can be found in Appendix~\ref{app:eval_dim_details}.

\subsubsection{Specialized Automated Model: {\tt \textbf{LoMTL}}}
This component implements a ternary classifier to evaluate tutor responses across the four considered dimensions. The models developed by participants in the BEA shared task provide a good starting point, as summarized by \citet{kochmar-etal-2025-findings}. However, upon closer inspection, we found that several teams relied on closed-source models, fully fine-tuned and LoRA-tuned models, and some even used multiple models or ensembles for each dimension, which makes these approaches difficult to scale across all dimensions.

Considering this, we propose a novel LoRA-based multi-task model (called \textbf{{\tt LoMTL}}) that fine-tunes a single small LLM in a LoRA setting across all four dimensions, each treated as a task in a multi-task learning setting. This modeling is naturally suited to the four tasks, all of which are ternary, closely related classification problems, allowing them to benefit from shared learning during training. Additionally, we use sampling and balanced batching methods to improve performance. We observe that a small {\tt google/gemma-2-2b-it} \cite{team2024gemma} model using {\tt LoMTL} achieves performance competitive with the top-performing teams in the shared task while remaining highly efficient and scalable. Appendix~\ref{app:lomtl_model} presents more details on model development, prompts, configuration, and comparative results with the BEA shared task's top-performing teams.

\subsubsection{LLM-as-a-Judge Evaluation}
This component provides functionality for evaluation using LLMs-as-judges. Following \citet{maurya-etal-2025-unifying}, we have selected {\tt Prometheus2}~\cite{kim-etal-2024-prometheus} as the primary open-source LLM. However, the implementation of this module is flexible enough to support the family of {\tt Llama} and other open-source causal LLMs for evaluation. Additionally, we have included an option to evaluate using the closed-source OpenAI {\tt GPT-5} model, although the implementation supports any model from the GPT family. Users must provide their OpenAI API key to run evaluations. We selected the above two models considering their open- and closed-source nature, as well as their human-like performance on public benchmarks~\cite{kim-etal-2024-prometheus}. However, the codebase is flexible enough to require only minimal adaptation to support other LLMs.

\subsubsection{Flexible Design}

Each of the three evaluation setups (automated evaluation, evaluation with open-source LLMs, and evaluation with closed-source GPT models) provides high degree of flexibility in the choice of the base LLM, prompting strategy, and hyper-parameter tuning. For example, each evaluation setup has its own prompting file and configuration, allowing users to customize and use the components of the tool as per their needs. Evaluation can be run using the following short commands:

\begin{tcolorbox}[
    colback=black!2,
    colframe=black!70,
    left=0.5pt,
]
\small 
\begin{verbatim}

# Training LoMTL model
cd src/autoeval && ./lora_finetune_runner.sh

# Evaluation with automated (i.e., LoMTL) model
cd src/autoeval && ./lora_evaluation_runner.sh

# Evaluation with open-source LLM
cd src/llmeval && 
python run_open_llm_as_judge_evaluation.py

# Evaluation with GPT5
cd src/llmeval && ./gpt5_eval_runner.sh
\end{verbatim}
\end{tcolorbox}


\subsection{Frontend Module: Demo App}
\label{sys:demo}


We present an interactive prototype built on top of \texttt{MRBench}, designed to evaluate the pedagogical abilities of AI tutors in educational dialogues. Users can explore the demo in two modes: (1) a \textbf{Static Mode}, where educators can access a deployed version containing 10 conversations from the \texttt{MRBench} test set evaluated by our fine-tuned model and share feedback on the helpfulness of tutor responses; and (2) a \textbf{Customized Mode}, where developers can run the interface locally to analyze their own datasets using either our {\tt LoMTL} model or their own one, supported by the provided code (the exact step-by-step details are supplied in the official GitHub README file\footnote{\url{https://github.com/kaushal0494/AITutor-EvalKit/blob/main/README.md}}). The UI consists of three key modules – \textbf{Automated Evaluation}, \textbf{LLM Evaluation}, and \textbf{Visualizer}, that together enable exploration, analysis, and visualization of the key aspects of educational dialogues.
\begin{itemize}
    \item \textbf{Automated Evaluation} provides automated evaluation results using our {\tt LoMTL} model on 4 evaluation dimensions. \vspace{-0.5em}
    \item \textbf{LLM Evaluation} leverages LLMs-as-judges for human-aligned evaluations. \vspace{-0.5em}
    \item \textbf{Visualizer} enables rich visual analytics for interpreting evaluation scores across 4 pedagogical dimensions on the {\tt MRBench} development set.
\end{itemize}
\vspace{-0.5em}

The UI supports the selection of evaluation methods and provides instant visualization of performance metrics through plots, bar charts, and spider graphs. 

\subsubsection{Automated Evaluation UI}

The Automated Evaluation module presents results generated by our {\tt LoMTL} model, assessing AI tutors across four pedagogical dimensions. As shown in Figure~\ref{fig:auto_overview}, users are first presented with an overview panel summarizing key statistics such as the number of topics, models, dimensions, and conversations. Users can evaluate a single tutor or compare two tutors side by side.

\textbf{Single Tutor Evaluation:} In this mode, users select a problem topic to view the complete student-tutor dialogue in the ``Context'' block (Figure~\ref{fig:context_block}). A drop-down menu allows selection of a tutor model, and the corresponding response is displayed in the ``Tutor Response'' block. Users can rate the usefulness of the response (\textit{Helpful}, \textit{To Some Extent}, or \textit{Not Helpful}) and optionally view the ground truth solution for better context. Upon clicking ``Get Auto-Evaluation Results,'' the system generates performance evaluations across the chosen dimensions, with results downloadable in a PNG, JPG, or JSON format. The ``Best Performance by Dimension'' panel highlights the top tutor(s) for each dimension, helping users quickly identify their pedagogical strengths.

\textbf{Tutor Comparison Mode:} This mode allows users to directly compare two tutors by enabling the ``Tutor Comparison Mode'' option. The selected tutors' responses are displayed side by side (Figure~\ref{fig:compare_response}), and users can provide quick feedback indicating which tutor performed better or mark both as equally good or bad. After choosing the evaluation dimensions and clicking ``Compare Tutor Responses,'' the interface presents a detailed two-column comparison of scores across all selected dimensions. To facilitate interpretation, the ``Comparison Visualization'' panel (Figure~\ref{fig:comparison_visualization}) provides four interactive views: the \textbf{Summary} view highlights the leading tutor for each evaluation dimension and identifies the overall winner; the \textbf{Spider Chart} offers a radar-style visualization comparing performance patterns across dimensions; the \textbf{Bar Chart} displays side-by-side scores for each dimension; and the \textbf{Differences} view illustrates the magnitude of score gaps between tutors. All visualizations can be exported in PNG or JPG formats for reporting or analysis. Finally, the ``Best Performance by Dimension'' panel summarizes the comparative strengths of the tutor pair, providing a concise overview of pedagogical differences. This mode supports structured, interpretable, and visually grounded benchmarking of AI tutor performance.

\subsubsection{LLM Evaluation UI}

The LLM Evaluation module enables advanced pedagogical assessment of AI tutor responses using LLMs as judges. It extends the functionality of the automated evaluation pipeline by leveraging LLMs to judge and compare tutor responses across four pedagogical dimensions. The UI supports three evaluation modes: \textit{single tutor evaluation}, \textit{tutor model comparison}, and \textit{LLM judge comparison}. As shown in Figure~\ref{fig:llm_eval_overview}, the overview panel summarizes available topics, conversations, tutor models, evaluation dimensions, and judge LLMs, providing a quick snapshot of the evaluation setup. Currently, two LLM judges are supported: \texttt{GPT-5} and \texttt{Prometheus-7B-v2.0}~\cite{kim-etal-2024-prometheus}.

\textbf{Single Tutor Evaluation:} This mode allows users to analyze how a selected tutor performs on a specific problem using an LLM as a judge. After selecting the problem, tutor, and LLM judge, users can generate dimension-wise evaluation results that reflect the LLM's assessment of the tutor's pedagogical performance. A ``Best Performance by Dimension'' panel highlights the top-performing tutor(s) for each dimension.

\textbf{Tutor Comparison Mode:} This mode enables side-by-side comparison of two tutor models on the same problem, judged by a selected LLM. Tutor responses are displayed together (Figure~\ref{fig:llm_tutor_compare_mode}), and upon comparison, the system presents dimension-wise evaluations and visualizations – such as Summary, Spider Chart, Bar Chart, and Differences – to clearly show relative strengths across pedagogical aspects.

\textbf{Judge Comparison Mode:} This mode compares how different LLM judges evaluate the same tutor response. The tutor's response is displayed once, and evaluations from both judges are presented in parallel with corresponding visualizations. This feature helps assess consistency between LLM judges and identify possible biases in their evaluations.

Together, these modes enable fine-grained, interpretable analysis of tutor behavior, offering insights into both model performance and evaluation reliability across different judging LLMs.

\subsubsection{Visualizer UI}

The Visualizer module provides a high-level overview of the \texttt{MRBench} development set, using gold-standard annotations across four evaluation dimensions through intuitive visual analytics. Users are first presented with a ``Dataset Overview'' panel summarizing key statistics, including the number of conversations, tutor models, and evaluation dimensions. This module includes three main visualization panels: {\em Tutor Performance Summary}, {\em Visualization Controls}, and {\em Dataset Visualization}.

The \textbf{Tutor Performance Summary} panel presents average scores for each tutor model across all four dimensions, where ``\emph{Yes},'' ``\emph{To some extent},'' and ``\emph{No}'' correspond to 1.0, 0.5, and 0.0, respectively (Figure~\ref{fig:tutor_summary}). The \textbf{Visualization Controls} panel allows users to select specific tutors and dimensions to generate detailed visualizations (Figure~\ref{fig:visualization_control}). The \textbf{Dataset Visualization} panel then displays the results through spider and bar charts, where spider plots (the \textit{most informative} visualizations from our perspective) summarize tutors' strengths and weaknesses across dimensions (Figure~\ref{fig:spider}), while bar charts show detailed score distributions for selected dimensions (Figure~\ref{fig:bar}), along with averages for each response label.

This visual analytics module enables users to explore and interpret pedagogical quality effectively, supporting comparative analysis and data-driven insights. The same functionality is also available in the customized mode, allowing developers to visualize and analyze their own datasets locally in a similar way.

\section{Evaluation}
\label{sec:eval}


To assess the toolkit and evaluation models, we first measure models' performance using the metrics from the BEA 2025 shared task~\cite{kochmar-etal-2025-findings} -- accuracy and macro-F1, and then conduct a human evaluation study, in which participants assess both the prediction quality of the \texttt{LoMTL} evaluation model and the usability of our demo tool and UI.

\subsection{Intrinsic Evaluation: Quantitative Analysis}
\label{eval:intrinsic}


For intrinsic evaluation, we compare our \texttt{LoMTL} model's predictions on the test set from \citet{kochmar-etal-2025-findings} with the gold human annotations and also run \texttt{GPT-5} and \texttt{Prometheus2} on the same data. The average results across evaluation dimensions, presented in \Cref{tab:intrinsic_results_average}, show that \texttt{Prometheus2} performs substantially worse than both our model and \texttt{GPT-5}. Our model achieves the highest averaged accuracy and macro-F1 on the full test set, and it also performs competitively on the demonstration subset. While \texttt{GPT-5} achieves a slightly higher averaged macro-F1 on the demonstration subset, the overall trend indicates that our model provides more reliable evaluations than \texttt{Prometheus2} and performs on par with, or better than, \texttt{GPT-5}. A close inspection of the confusion matrix between human annotations and the \texttt{LoMTL} model shows strong overall agreement, with the majority of instances concentrated on the diagonal, particularly for clear {\em Yes} (3,106) and {\em No} (1,057) cases. However, the model exhibits systematic confusion on borderline instances and a mild tendency to over-predict {\em Yes}, especially when humans assign {\em To some extent} or {\em No} labels. Extended results and a related discussion, including average precision and recall scores, are provided in Table~\ref{tab:intrinsic_results_extended} in Appendix~\ref{app:eval}. 

\begin{table}[t]
\centering
\small
\resizebox{\linewidth}{!}{
\begin{tabular}{lcccc}
\toprule
\textbf{Model} 
& \multicolumn{2}{c}{\textbf{Full Test Set}} 
& \multicolumn{2}{c}{\textbf{Demonstration Set}} \\
\cmidrule(lr){2-3} \cmidrule(lr){4-5}
& \textbf{Accuracy} & \textbf{Macro-F1}
& \textbf{Accuracy} & \textbf{Macro-F1} \\
\midrule
{\tt LoMTL} (ours) 
& \textbf{0.72} & \textbf{0.60}
& \textbf{0.68} & 0.55 \\

\texttt{Prometheus2} 
& 0.47 & 0.41
& 0.41 & 0.34 \\

\texttt{GPT-5} 
& 0.66 & 0.58
& 0.66 & \textbf{0.59} \\
\bottomrule
\end{tabular}
}
\caption{Accuracy and macro-F1 scores (averaged across Mistake Identification (MI), Mistake Location (ML), Providing Guidance (PG), and Actionability (AC)) for \texttt{LoMTL}, \texttt{Prometheus2}, and \texttt{GPT-5} on the full test set from \citet{kochmar-etal-2025-findings} and on the demonstration set. Best results are shown in \textbf{bold}.}
\label{tab:intrinsic_results_average}
\end{table}




\noindent\textbf{Performance across dimensions:} The per-dimension results, presented in \Cref{tab:intrinsic_results}, show that \texttt{Prometheus2} performs poorly, while our model notably outperforms \texttt{GPT-5} on the Mistake Identification and Actionability dimensions. However, it underperforms \texttt{GPT-5} by 3 and 6 percentage points in terms of macro-F1 for Mistake Location and Providing Guidance, respectively. A similar trend is observed in the ten dialogues used for demonstration.

\begin{table*}[t]
\centering
\small
\resizebox{\linewidth}{!}{
\begin{tabular}{lcccccccccccccccc}
\toprule
\textbf{Model} 
& \multicolumn{8}{c}{\textbf{Full Test Set}} 
& \multicolumn{8}{c}{\textbf{Demonstration Set}} \\
\cmidrule(lr){2-9} \cmidrule(lr){10-17}
 & \multicolumn{4}{c}{\textbf{Accuracy}} 
 & \multicolumn{4}{c}{\textbf{Macro-F1}} 
 & \multicolumn{4}{c}{\textbf{Accuracy}} 
 & \multicolumn{4}{c}{\textbf{Macro-F1}} \\
\cmidrule(lr){2-5} \cmidrule(lr){6-9} \cmidrule(lr){10-13} \cmidrule(lr){14-17}
 & \textbf{MI} & \textbf{ML} & \textbf{PG} & \textbf{AC} 
 & \textbf{MI} & \textbf{ML} & \textbf{PG} & \textbf{AC} 
 & \textbf{MI} & \textbf{ML} & \textbf{PG} & \textbf{AC} 
 & \textbf{MI} & \textbf{ML} & \textbf{PG} & \textbf{AC} \\
\midrule
{\tt LoMTL} (ours) & \textbf{0.86} & 0.67 & 0.63 & \textbf{0.70} & \textbf{0.66} & 0.55 & 0.54 & \textbf{0.65} & \textbf{0.76} & \textbf{0.69} & 0.60 & \textbf{0.68} & \textbf{0.57} & 0.50 & 0.47 & \textbf{0.65} \\
\texttt{Prometheus} & 0.58 & 0.53 & 0.31 & 0.46 & 0.48 & 0.42 & 0.32 & 0.43 & 0.48 & 0.41 & 0.42 & 0.32 & 0.34 & 0.30 & 0.38 & 0.30 \\
\texttt{GPT-5} & 0.67 & \textbf{0.68} & \textbf{0.70} & 0.58 & 0.53 & \textbf{0.58} & \textbf{0.61} & 0.55 & 0.67 & 0.64 & \textbf{0.67} & 0.66 & 0.53 & \textbf{0.56} & \textbf{0.59} & 0.64 \\
\bottomrule
\end{tabular}
}
\caption{Accuracy and macro-F1 scores of our model, \texttt{Prometheus}, and \texttt{GPT-5} across Mistake Identification (MI), Mistake Location (ML), Providing Guidance (PG), and Actionability (AC) on the full test set from \citet{kochmar-etal-2025-findings} and on the demonstration set. Best results are shown in \textbf{bold}.}
\label{tab:intrinsic_results}
\end{table*}

\subsection{Extrinsic Evaluation: Human Study}
\label{eval:extrinsic}

Participants were given access to the demo website, detailed guidelines, and an evaluation form. The form instructed them to assess three components: the Automated Evaluation tab, the LLM Evaluation tab, and the Visualizer tab, as detailed in Section \ref{sys:demo}. For the first two tabs, participants were asked to explore at least five different dialogues, review at least two tutor responses per dialogue, and use the feedback field to rate each response as \textit{helpful}, \textit{helpful to some extent}, or \textit{not helpful}. They also examined the model's evaluation for each response across at least one assessment dimension. Additionally, for each dialogue, participants compared at least one pair of tutor responses in comparison mode and indicated which response they considered a better one. After using Automated Evaluation tab, they reported how frequently they agreed with the model's judgments on a 1-5 scale, both in single-response and comparison modes. In the LLM Evaluation tab, participants evaluated how often they agreed with \texttt{GPT-5} and \texttt{Prometheus2}, again in both modes. They were also asked which evaluation model they perceived as more accurate: \texttt{GPT-5} or the model from the first tab (which corresponds to our \texttt{LoMTL} model), and \texttt{Prometheus2} or the model from the first tab. Finally, in the Visualizer tab, participants explored visualizations for at least two tutors and rated how informative they found them on a 1-5 scale. For every tab, participants also rated the ease of use on a 1-5 scale. The full questionnaire is provided in Appendix \ref{app:eval}.

A total of 14 participants took part in the study. Their educational background included individuals pursuing a Master's degree, those holding a Master's degree, and those holding a PhD. Eleven participants had teaching experience, and eight had prior experience using an AI tutor.

\begin{figure}[h]
    \centering
    \includegraphics[width=\linewidth]{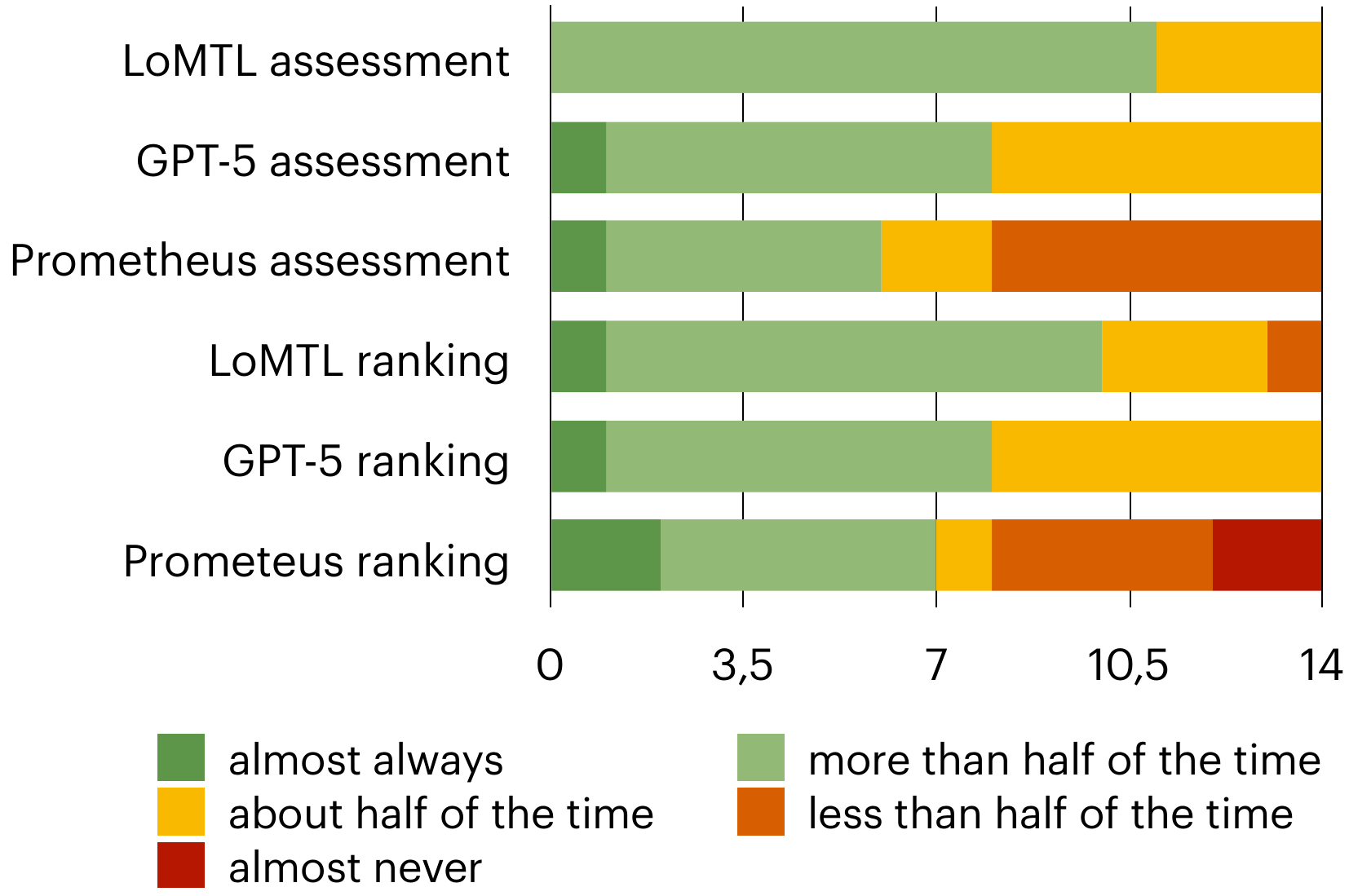}
    \caption{Participants' responses indicating how often they agreed with the models' judgments in single-response and comparison modes.}
    \label{fig:assessment_agreement}
\end{figure}

Most participants perceived the {\tt LoMTL} evaluation model as more accurate in its judgments than both \texttt{GPT-5} and \texttt{Prometheus2}. As shown in \Cref{fig:assessment_agreement}, the majority of participants agreed with the {\tt LoMTL} model's assessments more than half of the time in both single-response and comparison modes. A similar trend was observed for \texttt{GPT-5}: most participants agreed with its judgments more than half of the time, although a larger proportion reported agreement only about half of the time. In contrast, participants tended to agree with \texttt{Prometheus2} less than half of the time in the single-response mode, but more than half of the time in the comparison mode. Overall, most participants rated the ease of use of all tabs as \textit{very easy} and found the visualizations \textit{very informative}.

In total, we collected 95 annotations for single-tutor responses and 115 for pairwise comparisons. The analysis of these annotations is provided in Appendix~\ref{ap:annotation_analysis}. We do not present it here because evaluating tutor performance is not the focus of this work. Instead, our goal is to demonstrate how our interactive UI can be used for annotation purposes by collecting data that can later be used to train evaluation models or align language models.



\section{Conclusions and Future Work}
\label{sec:future}
This paper introduces the first open-access, open-source model for pedagogical quality evaluation of AI tutor responses, released under an MIT license. It is grounded on pedagogical principles and presents multiple evaluation choices including our proposed light-weight multi-tasking \texttt{LoMTL} model.  The toolkit is highly customizable, allowing researchers and practitioners to extend it to diverse educational contexts and dialogues across domains, and it can be easily set up and run locally. In addition, we provide an interactive web-based UI that offers interpretable evaluations of the pedagogical capabilities of state-of-the-art LLMs acting as AI tutors for education stakeholders, policymakers, and non-technical audience.

Future work will extend the toolkit by (1) integrating new user dialogues and LLMs in the frontend module to generate real-time responses and evaluations; (2) expanding the evaluation dimensions; (3) enabling new data upload option in UI; and (4) broadening coverage to additional subjects, grade levels, and languages.

\section*{Limitations}

We acknowledge that our work has several limitations. Below, we summarize the major ones among them.

\paragraph{Domain and grade level:} In this work, we focus on the mathematical tasks at the middle-school level. This decision is motivated by the availability of the data at this level and in this domain, but we plan to extend our work to other domains and levels as we elaborate in Section \ref{sec:future}. Moreover, since users can run our tool on their own data, new domains and levels can already be integrated on the user's side.
\paragraph{Language:} Similarly, our current work focuses on English only. In the future, we hope to extend it to other languages, as we specify in Section \ref{sec:future}, while users of our tool can also apply it to data in other languages with their own evaluation models and LLMs-as-judges on their side.
\paragraph{Educational scenario:} Building on previous work in this domain~\cite{maurya-etal-2025-unifying,kochmar-etal-2025-findings}, we only address student mistake remediation as an educational scenario. While this is one of the most salient and challenging scenarios in educational dialogues, we recognize this as one of the limitations of our work and plan to address it in the future.
\paragraph{Context length:} Similarly, following up on the previous work, our current evaluation approach is limited to a single turn in the dialogue. We acknowledge this as a limitation, and believe that future work should extend single-turn approaches to multiple-turn or full-dialogue ones.
\paragraph{Taxonomy:} We have built our prototype tool around a well-established evaluation taxonomy of~\citet{maurya-etal-2025-unifying}. While using a single taxonomy is a limitation, our code is open-source and can be extended to incorporate other data, dialogues, taxonomies, and evaluation models on the user's side.
\paragraph{Models:} Finally, via our demo tool, we only showcase a few LLMs-as-tutors and deploy only two LLMs-as-judges. Since our code is open-source and extendable, more tutor models and LLMs-as-judges can be integrated on the user's side.

\section*{Ethical Considerations}

As this work is exploratory, we do not anticipate any significant ethical risks associated with it. Moreover, one of our main goals in this research is to raise awareness of the education stakeholders as well as practitioners interested in the state of the art in AI in Education and the current capabilities of LLMs-as-tutors. Through transparent evaluation of such models, we aim to improve their interpretability, which we hope will help avoid potential future risks associated with a wider adoption of these models in education. 

This work uses the MRBench dataset \cite{maurya-etal-2025-unifying}, which integrates the MathDial \cite{macina-etal-2023-mathdial} and Bridge \cite{wang-etal-2024-bridging} datasets. In these datasets, the identities of the tutors are not revealed, while the student profiles are either synthetically created or anonymized. As a result, we do not anticipate any direct ethical risks associated with the datasets used. 

\section*{Acknowledgments}

We are grateful to the Google Academic Research Award (GARA) 2024 for supporting this research.


\bibliography{custom}

\begin{thebibliography}{36}
\providecommand{\natexlab}[1]{#1}

\bibitem[{Abdin et~al.(2024)Abdin, Jacobs, Awan, Aneja, Awadallah, Awadalla, Bach, Bahree, Bakhtiari, Behl et~al.}]{abdin2024phi}
Marah Abdin, Sam~Ade Jacobs, Ammar~Ahmad Awan, Jyoti Aneja, Ahmed Awadallah, Hany Awadalla, Nguyen Bach, Amit Bahree, Arash Bakhtiari, Harkirat Behl, and 1 others. 2024.
\newblock {Phi-3 technical report: A highly capable language model locally on your phone}.
\newblock \emph{arXiv preprint arXiv:2404.14219}.

\bibitem[{Achiam et~al.(2023)Achiam, Adler, Agarwal, Ahmad, Akkaya, Aleman, Almeida, Altenschmidt, Altman, Anadkat et~al.}]{achiam2023gpt}
Josh Achiam, Steven Adler, Sandhini Agarwal, Lama Ahmad, Ilge Akkaya, Florencia~Leoni Aleman, Diogo Almeida, Janko Altenschmidt, Sam Altman, Shyamal Anadkat, and 1 others. 2023.
\newblock {Gpt-4 technical report}.
\newblock \emph{arXiv preprint arXiv:2303.08774}.

\bibitem[{An et~al.(2025)An, Fu, Liu, Zong, Kong, Liu, Wang, Liu, Yang, Fan, and Yang}]{an-etal-2025-blcu}
Jiyuan An, Xiang Fu, Bo~Liu, Xuquan Zong, Cunliang Kong, Shuliang Liu, Shuo Wang, Zhenghao Liu, Liner Yang, Hanghang Fan, and Erhong Yang. 2025.
\newblock \href {https://doi.org/10.18653/v1/2025.bea-1.84} {{BLCU}-{ICALL} at {BEA} 2025 shared task: Multi-strategy evaluation of {AI} tutors}.
\newblock In \emph{Proceedings of the 20th Workshop on Innovative Use of NLP for Building Educational Applications (BEA 2025)}, pages 1084--1097, Vienna, Austria. Association for Computational Linguistics.

\bibitem[{Anthropic(2024)}]{TheC3}
Anthropic. 2024.
\newblock {The Claude 3 Model Family: Opus, Sonnet, Haiku}.
\newblock In \emph{\url{https://api.semanticscholar.org/CorpusID:268232499}}.

\bibitem[{Bird(2006)}]{bird2006nltk}
Steven Bird. 2006.
\newblock {NLTK: the natural language toolkit}.
\newblock In \emph{Proceedings of the COLING/ACL 2006 Interactive Presentation Sessions}, pages 69--72.

\bibitem[{Bloom(1984)}]{Bloom1984The2S}
Benjamin~Samuel Bloom. 1984.
\newblock \href {https://api.semanticscholar.org/CorpusID:1714225} {{The 2 Sigma Problem: The Search for Methods of Group Instruction as Effective as One-to-One Tutoring}}.
\newblock \emph{Educational Researcher}, 13:16 -- 4.

\bibitem[{Boaler(2013)}]{boaler2013ability}
Jo~Boaler. 2013.
\newblock {Ability and mathematics: The mindset revolution that is reshaping education}.
\newblock Forum.

\bibitem[{Boyd et~al.(2008)Boyd, Grossman, Lankford, Loeb, and Wyckoff}]{Boyd2008TeacherPA}
Donald~J. Boyd, Pam~L. Grossman, Hamilton Lankford, Susanna Loeb, and James~Humphrey Wyckoff. 2008.
\newblock \href {https://api.semanticscholar.org/CorpusID:40388601} {{Teacher Preparation and Student Achievement}}.
\newblock \emph{Educational Evaluation and Policy Analysis}, 31:416--440.

\bibitem[{Chang et~al.(2020)Chang, Chiam, Fu, Wang, Zhang, and Danescu-Niculescu-Mizil}]{chang2020convokit}
Jonathan~P Chang, Caleb Chiam, Liye Fu, Andrew~Z Wang, Justine Zhang, and Cristian Danescu-Niculescu-Mizil. 2020.
\newblock {Convokit: A toolkit for the analysis of conversations}.
\newblock \emph{arXiv preprint arXiv:2005.04246}.

\bibitem[{Dubey et~al.(2024)Dubey, Jauhri, Pandey, Kadian, Al-Dahle, Letman, Mathur, Schelten, Yang, Fan et~al.}]{dubey2024llama}
Abhimanyu Dubey, Abhinav Jauhri, Abhinav Pandey, Abhishek Kadian, Ahmad Al-Dahle, Aiesha Letman, Akhil Mathur, Alan Schelten, Amy Yang, Angela Fan, and 1 others. 2024.
\newblock {The LLaMA 3 herd of models}.
\newblock \emph{arXiv preprint arXiv:2407.21783}.

\bibitem[{Fan et~al.(2025)Fan, Tan, and Song}]{fan-etal-2025-bjtu}
Yuming Fan, Chuangchuang Tan, and Wenyu Song. 2025.
\newblock \href {https://doi.org/10.18653/v1/2025.bea-1.82} {{BJTU} at {BEA} 2025 shared task: Task-aware prompt tuning and data augmentation for evaluating {AI} math tutors}.
\newblock In \emph{Proceedings of the 20th Workshop on Innovative Use of NLP for Building Educational Applications (BEA 2025)}, pages 1073--1077, Vienna, Austria. Association for Computational Linguistics.

\bibitem[{Handa et~al.(2023)Handa, Clapper, Boyle, Wang, Yang, Yeager, and Demszky}]{handa-etal-2023-mistakes}
Kunal Handa, Margarett Clapper, Jessica Boyle, Rose Wang, Diyi Yang, David Yeager, and Dorottya Demszky. 2023.
\newblock \href {https://doi.org/10.18653/v1/2023.emnlp-main.549} {{{``}Mistakes Help Us Grow{''}: Facilitating and Evaluating Growth Mindset Supportive Language in Classrooms}}.
\newblock In \emph{Proceedings of the 2023 Conference on Empirical Methods in Natural Language Processing}, pages 8877--8897, Singapore. Association for Computational Linguistics.

\bibitem[{Hikal et~al.(2025)Hikal, Basem, Oshallah, and Hamdi}]{hikal-etal-2025-msa}
Baraa Hikal, Mohamed Basem, Islam Oshallah, and Ali Hamdi. 2025.
\newblock \href {https://doi.org/10.18653/v1/2025.bea-1.95} {{MSA} at {BEA} 2025 shared task: Disagreement-aware instruction tuning for multi-dimensional evaluation of {LLM}s as math tutors}.
\newblock In \emph{Proceedings of the 20th Workshop on Innovative Use of NLP for Building Educational Applications (BEA 2025)}, pages 1194--1202, Vienna, Austria. Association for Computational Linguistics.

\bibitem[{Jiang et~al.(2023)Jiang, Sablayrolles, Mensch, Bamford, Chaplot, Casas, Bressand, Lengyel, Lample, Saulnier et~al.}]{jiang2023mistral}
Albert~Q Jiang, Alexandre Sablayrolles, Arthur Mensch, Chris Bamford, Devendra~Singh Chaplot, Diego de~las Casas, Florian Bressand, Gianna Lengyel, Guillaume Lample, Lucile Saulnier, and 1 others. 2023.
\newblock {Mistral 7B}.
\newblock \emph{arXiv preprint arXiv:2310.06825}.

\bibitem[{Jurenka et~al.(2024)Jurenka, Kunesch, McKee, Gillick, Zhu, Wiltberger, Phal, Hermann, Kasenberg, Bhoopchand et~al.}]{jurenka2024towards}
Irina Jurenka, Markus Kunesch, Kevin~R McKee, Daniel Gillick, Shaojian Zhu, Sara Wiltberger, Shubham~Milind Phal, Katherine Hermann, Daniel Kasenberg, Avishkar Bhoopchand, and 1 others. 2024.
\newblock {Towards responsible development of generative AI for education: An evaluation-driven approach}.
\newblock \emph{arXiv preprint arXiv:2407.12687}.

\bibitem[{Kelly et~al.(2020)Kelly, Bringe, Aucejo, and Fruehwirth}]{Kelly2020UsingGO}
Sean Kelly, Robert Bringe, Esteban Aucejo, and Jane~Cooley Fruehwirth. 2020.
\newblock {Using global observation protocols to inform research on teaching effectiveness and school improvement: Strengths and emerging limitations}.
\newblock \emph{Education Policy Analysis Archives}, 28:62--62.

\bibitem[{Khan(2024)}]{khan2024khanmigo}
Sal Khan. 2024.
\newblock {Khanmigo}.

\bibitem[{Kim et~al.(2024)Kim, Suk, Longpre, Lin, Shin, Welleck, Neubig, Lee, Lee, and Seo}]{kim-etal-2024-prometheus}
Seungone Kim, Juyoung Suk, Shayne Longpre, Bill~Yuchen Lin, Jamin Shin, Sean Welleck, Graham Neubig, Moontae Lee, Kyungjae Lee, and Minjoon Seo. 2024.
\newblock \href {https://doi.org/10.18653/v1/2024.emnlp-main.248} {{Prometheus 2: An Open Source Language Model Specialized in Evaluating Other Language Models}}.
\newblock In \emph{Proceedings of the 2024 Conference on Empirical Methods in Natural Language Processing}, pages 4334--4353, Miami, Florida, USA. Association for Computational Linguistics.

\bibitem[{Kochmar et~al.(2025)Kochmar, Maurya, Petukhova, Srivatsa, Tack, and Vasselli}]{kochmar-etal-2025-findings}
Ekaterina Kochmar, Kaushal Maurya, Kseniia Petukhova, KV~Aditya Srivatsa, Ana{\"i}s Tack, and Justin Vasselli. 2025.
\newblock \href {https://doi.org/10.18653/v1/2025.bea-1.77} {Findings of the {BEA} 2025 shared task on pedagogical ability assessment of {AI}-powered tutors}.
\newblock In \emph{Proceedings of the 20th Workshop on Innovative Use of NLP for Building Educational Applications (BEA 2025)}, pages 1011--1033, Vienna, Austria. Association for Computational Linguistics.

\bibitem[{Kosmyna et~al.(2025)Kosmyna, Hauptmann, Yuan, Situ, Liao, Beresnitzky, Braunstein, and Maes}]{kosmyna2025your}
Nataliya Kosmyna, Eugene Hauptmann, Ye~Tong Yuan, Jessica Situ, Xian-Hao Liao, Ashly~Vivian Beresnitzky, Iris Braunstein, and Pattie Maes. 2025.
\newblock Your brain on chatgpt: Accumulation of cognitive debt when using an ai assistant for essay writing task.
\newblock \emph{arXiv preprint arXiv:2506.08872}.

\bibitem[{Macina et~al.(2023{\natexlab{a}})Macina, Daheim, Chowdhury, Sinha, Kapur, Gurevych, and Sachan}]{macina-etal-2023-mathdial}
Jakub Macina, Nico Daheim, Sankalan Chowdhury, Tanmay Sinha, Manu Kapur, Iryna Gurevych, and Mrinmaya Sachan. 2023{\natexlab{a}}.
\newblock \href {https://doi.org/10.18653/v1/2023.findings-emnlp.372} {{{M}ath{D}ial: A Dialogue Tutoring Dataset with Rich Pedagogical Properties Grounded in Math Reasoning Problems}}.
\newblock In \emph{Findings of the Association for Computational Linguistics: EMNLP 2023}, pages 5602--5621, Singapore. Association for Computational Linguistics.

\bibitem[{Macina et~al.(2025)Macina, Daheim, Hakimi, Kapur, Gurevych, and Sachan}]{macina-etal-2025-mathtutorbench}
Jakub Macina, Nico Daheim, Ido Hakimi, Manu Kapur, Iryna Gurevych, and Mrinmaya Sachan. 2025.
\newblock \href {https://doi.org/10.18653/v1/2025.emnlp-main.11} {{M}ath{T}utor{B}ench: A benchmark for measuring open-ended pedagogical capabilities of {LLM} tutors}.
\newblock In \emph{Proceedings of the 2025 Conference on Empirical Methods in Natural Language Processing}, pages 204--221, Suzhou, China. Association for Computational Linguistics.

\bibitem[{Macina et~al.(2023{\natexlab{b}})Macina, Daheim, Wang, Sinha, Kapur, Gurevych, and Sachan}]{macina-etal-2023-opportunities}
Jakub Macina, Nico Daheim, Lingzhi Wang, Tanmay Sinha, Manu Kapur, Iryna Gurevych, and Mrinmaya Sachan. 2023{\natexlab{b}}.
\newblock \href {https://doi.org/10.18653/v1/2023.eacl-main.173} {{Opportunities and Challenges in Neural Dialog Tutoring}}.
\newblock In \emph{Proceedings of the 17th Conference of the European Chapter of the Association for Computational Linguistics}, pages 2357--2372, Dubrovnik, Croatia. Association for Computational Linguistics.

\bibitem[{Mao et~al.(2025)Mao, Ge, Fan, Xu, Mi, Hu, and Gao}]{mao2025survey}
Yuren Mao, Yuhang Ge, Yijiang Fan, Wenyi Xu, Yu~Mi, Zhonghao Hu, and Yunjun Gao. 2025.
\newblock A survey on lora of large language models.
\newblock \emph{Frontiers of Computer Science}, 19(7):197605.

\bibitem[{Maurya et~al.(2025)Maurya, Srivatsa, Petukhova, and Kochmar}]{maurya-etal-2025-unifying}
Kaushal~Kumar Maurya, Kv~Aditya Srivatsa, Kseniia Petukhova, and Ekaterina Kochmar. 2025.
\newblock \href {https://doi.org/10.18653/v1/2025.naacl-long.57} {Unifying {AI} tutor evaluation: An evaluation taxonomy for pedagogical ability assessment of {LLM}-powered {AI} tutors}.
\newblock In \emph{Proceedings of the 2025 Conference of the Nations of the Americas Chapter of the Association for Computational Linguistics: Human Language Technologies (Volume 1: Long Papers)}, pages 1234--1251, Albuquerque, New Mexico. Association for Computational Linguistics.

\bibitem[{Minaee et~al.(2024)Minaee, Mikolov, Nikzad, Chenaghlu, Socher, Amatriain, and Gao}]{minaee2024large}
Shervin Minaee, Tomas Mikolov, Narjes Nikzad, Meysam Chenaghlu, Richard Socher, Xavier Amatriain, and Jianfeng Gao. 2024.
\newblock {Large language models: A survey}.
\newblock \emph{arXiv preprint arXiv:2402.06196}.

\bibitem[{Pedregosa et~al.(2011)Pedregosa, Varoquaux, Gramfort, Michel, Thirion, Grisel, Blondel, Prettenhofer, Weiss, Dubourg, Vanderplas, Passos, Cournapeau, Brucher, Perrot, and Duchesnay}]{pedregosa2011scikit}
Fabian Pedregosa, Ga{\"e}l Varoquaux, Alexandre Gramfort, Vincent Michel, Bertrand Thirion, Olivier Grisel, Mathieu Blondel, Peter Prettenhofer, Ron Weiss, Vincent Dubourg, Jake Vanderplas, Alexandre Passos, David Cournapeau, Mathieu Brucher, Matthieu Perrot, and {\'E}douard Duchesnay. 2011.
\newblock {Scikit-learn: Machine Learning in {P}ython}.
\newblock \emph{Journal of Machine Learning Research}, 12:2825--2830.

\bibitem[{Reid et~al.(2024)Reid, Savinov, Teplyashin, Lepikhin, Lillicrap, Alayrac, Soricut, Lazaridou, Firat, Schrittwieser et~al.}]{reid2024gemini}
Machel Reid, Nikolay Savinov, Denis Teplyashin, Dmitry Lepikhin, Timothy Lillicrap, Jean-baptiste Alayrac, Radu Soricut, Angeliki Lazaridou, Orhan Firat, Julian Schrittwieser, and 1 others. 2024.
\newblock {Gemini 1.5: Unlocking multimodal understanding across millions of tokens of context}.
\newblock \emph{arXiv preprint arXiv:2403.05530}.

\bibitem[{Roh and Bang(2025)}]{roh-bang-2025-bea}
Jihyeon Roh and Jinhyun Bang. 2025.
\newblock \href {https://doi.org/10.18653/v1/2025.bea-1.80} {bea-jh at {BEA} 2025 shared task: Evaluating {AI}-powered tutors through pedagogically-informed reasoning}.
\newblock In \emph{Proceedings of the 20th Workshop on Innovative Use of NLP for Building Educational Applications (BEA 2025)}, pages 1049--1059, Vienna, Austria. Association for Computational Linguistics.

\bibitem[{Tack et~al.(2023)Tack, Kochmar, Yuan, Bibauw, and Piech}]{tack-etal-2023-bea}
Ana{\"i}s Tack, Ekaterina Kochmar, Zheng Yuan, Serge Bibauw, and Chris Piech. 2023.
\newblock \href {https://doi.org/10.18653/v1/2023.bea-1.64} {The {BEA} 2023 shared task on generating {AI} teacher responses in educational dialogues}.
\newblock In \emph{Proceedings of the 18th Workshop on Innovative Use of NLP for Building Educational Applications (BEA 2023)}, pages 785--795, Toronto, Canada. Association for Computational Linguistics.

\bibitem[{Team et~al.(2024)Team, Riviere, Pathak, Sessa, Hardin, Bhupatiraju, Hussenot, Mesnard, Shahriari, Ram{\'e} et~al.}]{team2024gemma}
Gemma Team, Morgane Riviere, Shreya Pathak, Pier~Giuseppe Sessa, Cassidy Hardin, Surya Bhupatiraju, L{\'e}onard Hussenot, Thomas Mesnard, Bobak Shahriari, Alexandre Ram{\'e}, and 1 others. 2024.
\newblock Gemma 2: Improving open language models at a practical size.
\newblock \emph{arXiv preprint arXiv:2408.00118}.

\bibitem[{Wang and Demszky(2024)}]{wang-demszky-2024-edu}
Rose Wang and Dorottya Demszky. 2024.
\newblock \href {https://doi.org/10.18653/v1/2024.naacl-demo.6} {{Edu-{C}onvo{K}it: An Open-Source Library for Education Conversation Data}}.
\newblock In \emph{Proceedings of the 2024 Conference of the North American Chapter of the Association for Computational Linguistics: Human Language Technologies (Volume 3: System Demonstrations)}, pages 61--69, Mexico City, Mexico. Association for Computational Linguistics.

\bibitem[{Wang et~al.(2024{\natexlab{a}})Wang, Zhang, Robinson, Loeb, and Demszky}]{wang-etal-2024-bridging}
Rose Wang, Qingyang Zhang, Carly Robinson, Susanna Loeb, and Dorottya Demszky. 2024{\natexlab{a}}.
\newblock \href {https://doi.org/10.18653/v1/2024.naacl-long.120} {{Bridging the Novice-Expert Gap via Models of Decision-Making: A Case Study on Remediating Math Mistakes}}.
\newblock In \emph{Proceedings of the 2024 Conference of the North American Chapter of the Association for Computational Linguistics: Human Language Technologies (Volume 1: Long Papers)}, pages 2174--2199, Mexico City, Mexico. Association for Computational Linguistics.

\bibitem[{Wang et~al.(2024{\natexlab{b}})Wang, Ribeiro, Robinson, Loeb, and Demszky}]{wang2024tutor}
Rose~E Wang, Ana~T Ribeiro, Carly~D Robinson, Susanna Loeb, and Dorottya Demszky. 2024{\natexlab{b}}.
\newblock {Tutor CoPilot: A Human-AI Approach for Scaling Real-Time Expertise. EdWorkingPaper No. 24-1054.}
\newblock \emph{Annenberg Institute for School Reform at Brown University}.

\bibitem[{Wolf(2019)}]{wolf2019huggingface}
T~Wolf. 2019.
\newblock {Huggingface's transformers: State-of-the-art natural language processing}.
\newblock \emph{arXiv preprint arXiv:1910.03771}.

\bibitem[{Yoon et~al.(2007)Yoon, Duncan, Lee, Scarloss, and Shapley}]{Yoon2007ReviewingTE}
Kwang~Suk Yoon, Teresa Duncan, Silvia Wen-Yu Lee, Beth Scarloss, and Kathy~L Shapley. 2007.
\newblock {Reviewing the evidence on how teacher professional development affects student achievement. issues \& answers. rel 2007-no. 033.}
\newblock ERIC.

\end{thebibliography}

\appendix

\clearpage
\section{Pedagogical Dimensions and Data}
\label{app:eval_dim_details}


\subsection{Evaluation Dimensions}
\citet{maurya-etal-2025-unifying} proposed an evaluation taxonomy with eight dimensions to assess the pedagogical soundness of \(T_{t+1}\) tutor response in the context of SMR. These dimensions are grounded in learning science research and prior work on tutor evaluation, defining the assessment via eight concrete criteria. Furthermore, the authors validated these dimensions as necessary and sufficient through a human pilot study. They also released the associated {\tt MRBench} dataset with human annotations (see $\S$\ref{app:dataset-details-stat}). Following this, \citet{kochmar-etal-2025-findings} focused on four key dimensions from this taxonomy for the BEA shared task, where participating teams were challenged to develop a ternary classification model for each of the four dimensions. These classifiers aimed to assess the pedagogical quality of the \(T_{t+1}\) response \textit{on-the-fly}, enabling evaluation scalability with new data and tutors. Details on each of the four dimensions, along with their definitions, annotated labels, and desiderata, are provided in Table~\ref{tab:detailed_def}. Building on this, our \texttt{AITutor-EvalKit} toolkit focuses on the four key dimensions used in the BEA shared task.

\begin{table*}[!htb]
    \centering
    \resizebox{\textwidth}{!}{
    \begin{tabular}{|l|l|l|c|}
         \hline
        \textbf{Dimension} &  \textbf{Definition} & \textbf{Labels} & \textbf{Desiderata} \\ \hline
        Mistake identification & Has the tutor identified a mistake in a student’s response? & \makecell[l]{(1) Yes \\ (2) To some extent \\ (3) No} & Yes \\ \hline
        Mistake location & Does the tutor's response accurately point to a genuine mistake and its location? & \makecell[l]{(1) Yes \\ (2) To some extent \\ (3) No} & Yes \\ \hline
        Providing guidance & \makecell[l]{Does the tutor offer correct and relevant guidance, such as an explanation, \\ elaboration, hint, examples, and so on?} & \makecell[l]{(1) Yes \\ (2) To some extent \\ (3) No} & Yes \\ \hline
        Actionability & Is it clear from the tutor's feedback what the student should do next? & \makecell[l]{(1) Yes \\ (2) To some extent \\ (3) No} & Yes \\ \hline
    \end{tabular}}
    \caption{An overview of the evaluation taxonomy, associated definitions, annotation labels, and desired labels from \citet{maurya-etal-2025-unifying}.}
    \label{tab:detailed_def}
\end{table*}

\subsection{Dataset Details and Statistics}
\label{app:dataset-details-stat}
We used the \texttt{MRBench} dataset to develop the toolkit and the automated evaluation model (i.e., {\tt LoMTL}). The initial version of \texttt{MRBench}, released by \citet{maurya-etal-2025-unifying}, contains 191 dialogues. This was extended by \citet{kochmar-etal-2025-findings} which results in 300 dialogues in the development set and 191 dialogues in the test set. In this work, we use the extended version of the \texttt{MRBench} dataset.

The dataset is built on top of two public datasets -- {\tt MathDial} \cite{macina-etal-2023-mathdial} and {\tt Bridge} \cite{wang-etal-2024-bridging} -- which provide partial conversational histories from secondary and primary school-level mathematics, respectively, along with human tutor responses as \(T_{t+1}\). {\tt MathDial} includes only one expert tutor response, whereas Bridge includes two responses from expert and novice human tutors. Additionally, each dialogue includes seven \(T_{t+1}\) responses generated by seven \textit{state-of-the-art} LLMs-as-tutors, including GPT-4 \cite{achiam2023gpt}, Gemini \cite{reid2024gemini}, Sonnet \cite{TheC3}, Mistral \cite{jiang2023mistral}, Llama-3.1-8B and Llama-3.1-405B \cite{dubey2024llama}, and Phi3 \cite{abdin2024phi}.

All \(T_{t+1}\) responses (whether from human tutors or LLMs) are annotated by human annotators across four selected dimensions with labels "{\em Yes}," "{\em To some extent}," and "{\em No}," as detailed in Table~\ref{tab:detailed_def}. The proposed {\tt LoMTL} model is trained on the development set and evaluated on the test set; all results presented in this work are reported on the test set. We further split the development set into a 9:1 ratio for training and validation when developing the {\tt LoMTL} model. All model checkpoints were selected based on validation performance. Finally, we randomly selected a subset of 10 dialogues from the test set as a demonstration set, which is used in the demo app. These details are summarized in Table \ref{tab:mrbench_stats}.

\begin{table*}[!htb]
\centering
\small
\begin{tabular}{l l}
\hline
\textbf{Parameters} & \textbf{Value/Details} \\
\hline
Number of dialogues (Dev / Test / Total) & 300 \;/\; 191 \;/\; 491 \\
Number of tutor responses (Dev / Test / Total) & 2,476 \;/\; 1,547 \;/\; 4,023 \\
Number of tutors (Total) & 9 \\
Number of human tutors & 2 (1 expert, 1 novice) \\
Number of LLM tutors & 7 \\
LLM tutor models & GPT-4, Gemini, Sonnet, Mistral, \\
& Llama-3.1-8B, Llama-3.1-405B, Phi-3 \\
Source datasets & \texttt{MathDial}, \texttt{Bridge} \\
{\tt MathDial} & Expert human tutor only \\
{\tt Bridge}  & Expert and Novice human tutors\\
Demonstration set size & 10 dialogues (from test set) \\
\hline
\end{tabular}
\caption{Key details of the extended \texttt{MRBench} dataset.}
\label{tab:mrbench_stats}
\end{table*}

\section{{\tt \textbf{LoMTL}} Evaluation Model}
\label{app:lomtl_model}


\subsection{Motivation}
Building {\tt LoMTL} has a two-fold motivation: (1) The current human annotation-based pedagogical ability assessments presented by \citet{maurya-etal-2025-unifying} are static in nature. They are not scalable to new LLMs and tutoring systems, which are being developed very frequently nowadays. We need a reliable \textit{automated} evaluation model that can provide pedagogical assessment \textit{on-the-fly} for new tutors or responses and help track the progress of AI tutor abilities. (2) The BEA shared task \cite{kochmar-etal-2025-findings} is a good starting point for developing an automated evaluation model. More than 50 international teams participated in the challenge and proposed several novel modeling approaches, including diverse prompting strategies, full instruction tuning, LoRA-based finetuning, supervised finetuning, data augmentation, label balancing, ensembling and so on. However, most teams that participated in all four tracks did not develop a unified approach (with the exception of the MSA team \cite{hikal-etal-2025-msa}) and instead used models with a large number of parameters. This hinders the adaptability of these approaches, as their deployment becomes challenging and costly. 

These limitations motivated us to develop {\tt LoMTL}, a lightweight model with only 2 billion parameters, created by training the \texttt{google/gemma-2-2b-it} model using LoRA in a multi-task learning setting. It achieves comparable performance while being significantly more efficient (see Table \ref{tab:autocomp} for comparison). For instance, the top-performing BJTU team \cite{fan-etal-2025-bjtu} achieved a macro-F1 score of 0.645 using 288 billion parameters (across all four dimensions). In contrast, the {\tt LoMTL} model achieved 0.60 with only 2 billion parameters, approximately 0.7\% of the BJTU model's parameter count.

\begin{table*}[!htb]
\centering
\resizebox{\textwidth}{!}{
\begin{tabular}{lcccc}
\hline
\textbf{Team/Model} & \textbf{Macro-F1} & \textbf{\# LLMs} & \textbf{\# Parameters} & \textbf{Parameter Size vs.\ 2B} \\
\hline
BJTU \cite {fan-etal-2025-bjtu} & 0.646 & 4 ({\tt Qwen/Qwen2.5-72B}) & $4 \times 72\text{B} = 288\text{B}$ & $144\times$ larger \\
MSA \cite{hikal-etal-2025-msa} & 0.643 & $5 \times 4$ ({\tt mistralai/Mistral-7B-v0.1}) & $20 \times 7\text{B} = 140\text{B}$ & $70\times$ larger \\
BLCU-ICALL \cite{an-etal-2025-blcu} & 0.632 & 4 ({\tt Qwen/Qwen2.5-7B}) & $4 \times 7\text{B} = 28\text{B}$ & $14\times$ larger \\
bea-jh \cite{roh-bang-2025-bea} & 0.625 & 4 ({\tt zai-org/glm-4-9b}) & $4 \times 9\text{B} = 36\text{B}$ & $18\times$ larger \\ \hline
{\tt Prometheus2} \cite{kim-etal-2024-prometheus} & 0.410 & 4 ({\tt prometheus-7b-v2.0}) & $4 \times 7\text{B} = 28\text{B}$ & $14\times$ larger \\
{\tt GPT-5} \cite{achiam2023gpt} & 0.581 & - & - & - \\ \hline
{\tt LoMTL} (ours) & 0.601 & 1 ({\tt google/gemma-2-2b-it}) & $2\text{B}$ & - \\
\hline
\end{tabular}}
\caption{Comparison of macro-F1 scores and model parameters between our automated evaluation model (i.e., {\tt LoMTL}) and the top-performing teams in the BEA shared task and LLM-as-a-judge models across four dimensions (aka. tasks). Note that, (1) for a fair comparison, we include only the teams that participated in all four tracks of the shared task, and (2) since GPT-5 is a closed-source model, its parameter details are not publicly available.}
\label{tab:autocomp}
\end{table*}

\subsection{Training and Inference}
In this section, we provide details on the training and evaluation of the {\tt LoMTL} model. Inspired by the success of LoRA-based modeling from the BEA shared task \cite{kochmar-etal-2025-findings} and by the community \cite{mao2025survey}, we adapted a LoRA-based fine-tuning approach. Since we have a small amount of training data (approximately 2500 examples for each dimension) and the different tasks are somewhat related, the natural modeling choice is multi-task learning, where each evaluation dimension is formulated as a task. This LoRA-based multi-task fine-tuning approach (called {\tt LoMTL}) resulted in a compact single model and enabled flexibility in model deployment. Further, we experimented with a small {\tt google/gemma-2-2b-it} model, which resulted in fast inference. We observed two major issues during training: \textit{task imbalance} and \textit{label imbalance}. To mitigate task imbalance, we implemented a balance batching where each batch has an uniform number of examples from each task. For label imbalance, we explored several approaches such as focal loss, label sampling, loss weighting, and sampling methods. We obtained the best performance with oversampling where we randomly sample underrepresented examples in the training dataset. Model training and inference were done with a single 48GB A6000 GPU. The best checkpoints were obtained using validation data (10\% of the development dataset).

\subsection{Prompts and Configurations}
\begin{figure*}[!htb]
    \centering
    \includegraphics[width=0.85\linewidth]{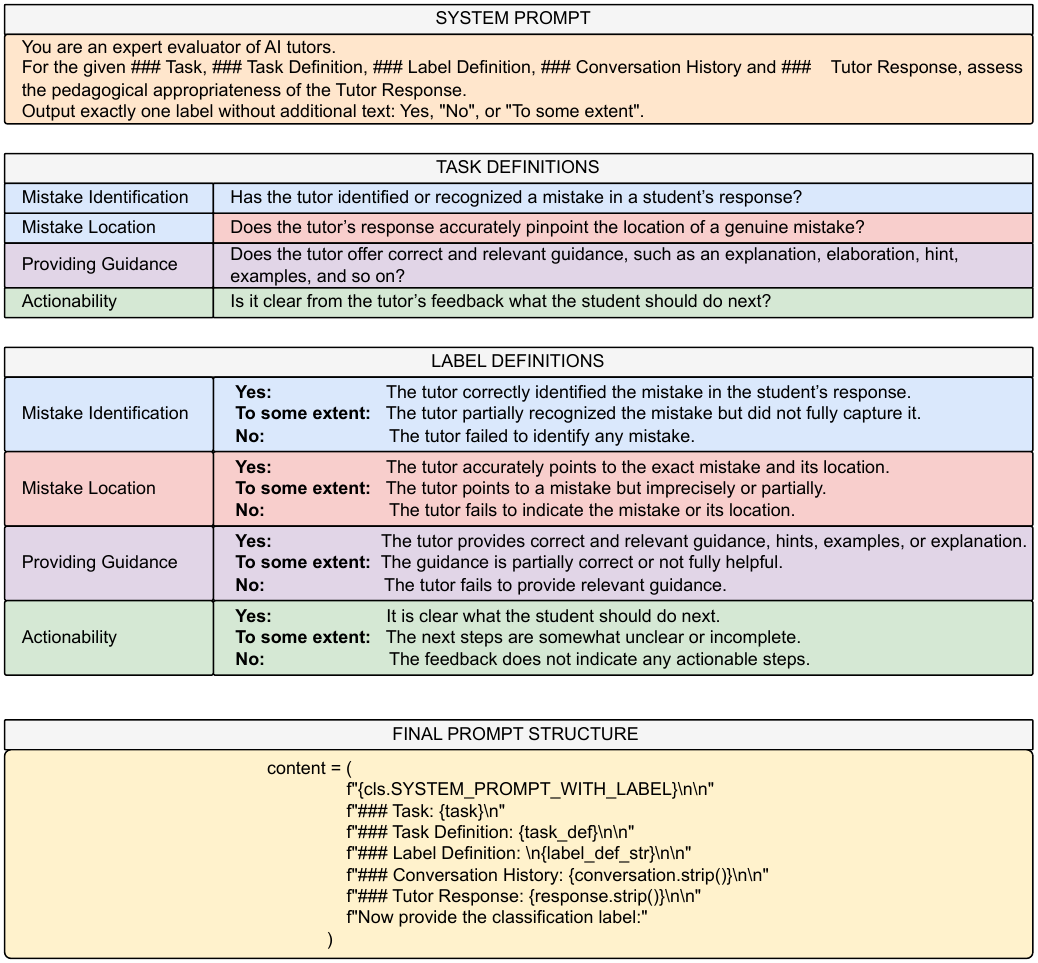}
    \caption{Overview of the prompt components, their associated definitions and details, and the final prompt structure used in {\tt LoMTL}.}
    \label{fig:promptlomtl}
\end{figure*}

Prompt structure and training/evaluation configurations for the {\tt LoMTL} model are shown in Figure \ref{fig:promptlomtl} and Table \ref{tab:paramsum}, respectively.

\begin{table*}[!htb]
\centering
\resizebox{0.8\textwidth}{!}{
\begin{tabular}{llp{7cm}}
\hline
\textbf{Category} & \textbf{Parameter} & \textbf{Value} \\
\hline

\multicolumn{3}{c}{\textbf{Common Settings}} \\
\hline
Model & MODEL\_NAME & {\tt google/gemma-2-2b-it} \\
Task Dimensions & DIMENSIONS & Mistake\_Identification, Mistake\_Location, Providing\_Guidance, Actionability \\
Input Length & MAX\_LENGTH & 1024 \\
Prompt & include\_label\_definitions & Enabled \\
\hline

\multicolumn{3}{c}{\textbf{Training-Only Settings}} \\
\hline
Batching & BATCH\_SIZE & 4 \\
Batching & GRAD\_ACCUM & 1 \\
Training Schedule & EPOCHS & 3 \\
Training Schedule & LEARNING\_RATE & 1e-4 \\
Training Schedule & WEIGHT\_DECAY & 0.1 \\
Logging & LOGGING\_STEPS & 50 \\
Saving & SAVE\_STEPS & 300 \\
Evaluation Cycle & EVAL\_STEPS & 300 \\
Oversampling & OVERSAMPLE\_METHOD & ``random'' \\
Metric for best model & METRIC\_FOR\_BEST & ``eval\_loss" \\
LoRA & LORA\_R & 8 \\
LoRA & LORA\_ALPHA & 16 \\
LoRA & LORA\_DROPOUT & 0.1 \\
Early Stopping & EARLY\_PATIENCE & 5 \\
Early Stopping & EARLY\_THRESHOLD & 0.0 \\
\hline

\multicolumn{3}{c}{\textbf{Evaluation-Only Settings}} \\
\hline
Generation & TEMPERATURE & 1.0 \\
\hline
\end{tabular}}
\caption{Summary of training and evaluation configurations along with their corresponding parameter names.}
\label{tab:paramsum}
\end{table*}

\section{Toolkit Evaluation Details and Results}
\label{app:eval}

\subsection{Extended Evaluation Results}

In addition to the observations on the accuracy and macro-F1 scores in Section \ref{eval:intrinsic}, a closer inspection of precision and recall (from Table \ref{tab:intrinsic_results_extended}) further supports our findings. On the full test set, {\tt LoMTL} achieves the highest macro-precision (0.63) while maintaining competitive recall (0.59), indicating more accurate and consistent positive predictions compared to both baselines. {\tt GPT-5} attains slightly higher recall (0.60) but with lower precision (0.58), suggesting a more recall-oriented behavior. On the demonstration subset, {\tt GPT-5} achieves the highest precision (0.60) and recall (0.61), whereas {\tt LoMTL} remains competitive (0.58 precision, 0.56 recall) and substantially outperforms {\tt Prometheus2}. Overall, these results show that LoMTL maintains a balanced precision–recall trade-off, reinforcing its robustness across evaluation dimensions.

\subsection{Human Annotation Analysis}
\label{ap:annotation_analysis}

We collected 95 annotations for single-tutor responses and 115 for pairwise comparisons. Since annotators were free to choose any dialogues and models, the number of annotations per tutor is not uniform. In pairwise comparisons, \texttt{Gemini} was selected most frequently -- 49 out of 115 comparisons included \texttt{Gemini} as one of the tutors. Other tutors appeared in 17-36 comparisons, except for Novice, which was chosen only five times. Overall, Expert responses were preferred most often, winning in 60\% of the cases in which the Expert's response appeared. \texttt{Sonnet} and \texttt{Llama-3.1-405B} were also selected more than half of the time. Novice's responses never won.

In the single-tutor mode, 37 annotations marked responses as \textit{helpful}, 31 as \textit{not helpful}, and 27 as \textit{to some extent helpful}. The most helpful responses came from Expert, \texttt{Sonnet}, \texttt{Gemini}, and \texttt{Mistral}, each of which was rated \textit{helpful} in at least half of the corresponding cases. The least helpful responses were from \texttt{Phi3} and \texttt{Llama-3.1-8B}. \texttt{Mistral} and \texttt{Llama-3.1-405B} were the only tutors without any \textit{not helpful} annotations, as they had the highest proportion of \textit{to some extent helpful} ratings.

\begin{table*}[!htb]
\centering
\small
\resizebox{\linewidth}{!}{
\begin{tabular}{lcccccccc}
\toprule
\textbf{Model} 
& \multicolumn{4}{c}{\textbf{Full Test Set}} 
& \multicolumn{4}{c}{\textbf{Demonstration Set}} \\
\cmidrule(lr){2-5} \cmidrule(lr){6-9}
& \textbf{Accuracy} & \textbf{Macro-F1} & \textbf{Precision} & \textbf{Recall}
& \textbf{Accuracy} & \textbf{Macro-F1} & \textbf{Precision} & \textbf{Recall}\\
\midrule
{\tt LoMTL} (ours) 
& \textbf{0.72} & \textbf{0.60} & \textbf{0.63} & 0.59
& \textbf{0.68} & 0.55 & 0.58 & 0.56\\

\texttt{Prometheus2} 
& 0.47 & 0.41 & 0.44 & 0.45
& 0.41 & 0.34 & 0.38 & 0.36\\

\texttt{GPT-5} 
& 0.66 & 0.58 & 0.58 & \textbf{0.60}
& 0.66 & \textbf{0.59} & \textbf{0.60 }& \textbf{0.61} \\
\bottomrule
\end{tabular}
}
\caption{Accuracy, macro-F1, macro-precision and macro-recall scores (averaged across Mistake Identification (MI), Mistake Location (ML), Providing Guidance (PG), and Actionability (AC)) for our \texttt{LoMTL} model, \texttt{Prometheus2}, and \texttt{GPT-5} on the full test set from \citet{kochmar-etal-2025-findings} and on the demonstration set. Best results are shown in \textbf{bold}.}
\label{tab:intrinsic_results_extended}
\end{table*}

\begin{figure*}[h]
    \centering
    \includegraphics[width=0.7\textwidth]{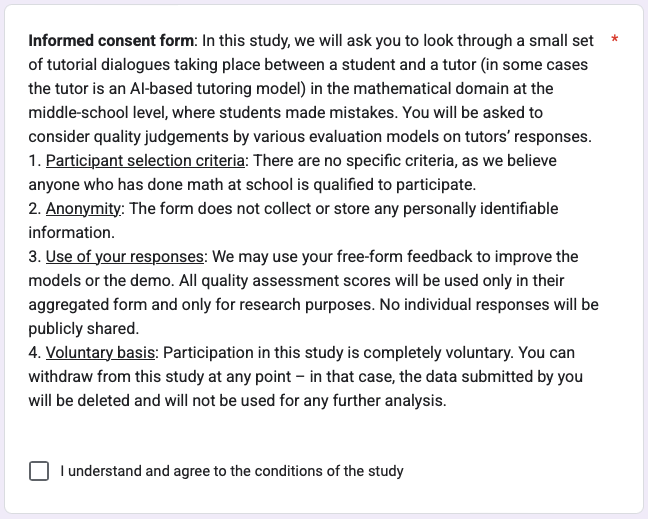}
    \caption{Informed consent form that participants were required to accept before proceeding with their feedback and annotations.}
    \label{fig:consent_form}
\end{figure*}

\onecolumn
\subsection{Full Questionnaire}

Below are the informed consent forms that participants were required to read and accept, as well as the full questionnaire.

\small
\setlength{\tabcolsep}{4pt}
\renewcommand{\arraystretch}{1.2}

\subsubsection*{Background}

\begin{longtable}{@{}p{0.40\textwidth}p{0.55\textwidth}@{}}
\toprule
\textbf{Item} & \textbf{Question / Response Options} \\
\midrule
\endhead

What is your highest qualification? &
Response options: \textit{Bachelor's degree; Master's degree; PhD degree; Other.} \\

Do you have teaching experience (e.g., lecturing, supervising or mentoring students, TA-ing, or similar)? &
Response options: \textit{Yes; No.} \\

Have you ever used an AI tutor before? &
Response options: \textit{Yes; No.} \\

\bottomrule
\end{longtable}

\subsubsection*{Automated Evaluation}

\begin{longtable}{@{}p{0.40\textwidth}p{0.55\textwidth}@{}}
\toprule
\textbf{Item} & \textbf{Question / Response Options} \\
\midrule
\endhead

Instructions &
This tab gives you an opportunity to select among 10 short dialogues on relatively simple (no higher than middle-school level) math problems, check students' misconceptions, and explore various human and AI tutor responses. The quality of these responses is evaluated using a fine-tuned evaluation model. \\

Steps &
1) Explore at least 5 different dialogues (the correct answer is also available to help you spot the student's mistake).\\
& 2) For each dialogue, check at least 2 different tutor responses.\\
& 3) Rate each response as \emph{Helpful}, \emph{Not Helpful}, or \emph{To some extent}.\\
& 4) Select at least one quality dimension to view the model's assessment of that response.\\
& 5) For each dialogue, use the comparison mode at least once. \\

How many dialogues have you checked? &
Free-form response. \\

How often did you agree with this model's assessment for a single tutor response? &
Scale: 1 = almost never, 2 = less than half of the time, 3 = about half of the time, 4 = more than half of the time, 5 = almost always.\\
Response options: \textit{1, 2, 3, 4, 5.} \\

How often did you agree with this model's ranking of the tutors in the comparison mode? &
Same scale as above.\\
Response options: \textit{1, 2, 3, 4, 5.} \\

On a scale from 1 to 5, how easy was it to use the ``Automated Evaluation'' tab? &
Scale: 1 = not easy at all, 5 = very easy.\\
Response options: \textit{1, 2, 3, 4, 5.} \\

Please feel free to give us any further feedback on this tab. &
Free-form response. \\

\bottomrule
\end{longtable}

\subsubsection*{LLM Evaluation}

\begin{longtable}{@{}p{0.40\textwidth}p{0.55\textwidth}@{}}
\toprule
\textbf{Item} & \textbf{Question / Response Options} \\
\midrule
\endhead

Instructions &
This tab provides access to the same dialogues and tutor responses as in Tab~1, but this time they are evaluated using LLMs as judges. You can choose between GPT-5 and Prometheus. \\

Steps &
1) Explore at least 5 different dialogues (they may be the same ones as before).\\
& 2) For each dialogue, check at least 2 different tutor responses.\\
& 3) Use each LLM-as-judge at least twice on different dialogues.\\
& 4) Select at least one quality dimension for each response.\\
& 5) For each dialogue, use the comparison mode at least once. \\

How many dialogues have you checked? &
Free-form response. \\

How often did you agree with GPT-5's assessment for a single tutor response? &
Scale: 1 = almost never, 2 = less than half of the time, 3 = about half of the time, 4 = more than half of the time, 5 = almost always.\\
Response options: \textit{1, 2, 3, 4, 5.} \\

How often did you agree with GPT-5's ranking of the tutors in the comparison mode? &
Same scale as above.\\
Response options: \textit{1, 2, 3, 4, 5.} \\

How often did you agree with Prometheus' assessment for a single tutor response? &
Same scale as above.\\
Response options: \textit{1, 2, 3, 4, 5.} \\

How often did you agree with Prometheus' ranking of the tutors in the comparison mode? &
Same scale as above.\\
Response options: \textit{1, 2, 3, 4, 5.} \\

Which evaluation model did you perceive to be more accurate in its judgments of tutor responses – GPT-5 or the fine-tuned model from Tab~1 (``Automated Evaluation'')? &
Response options: \textit{The model from Tab~1; GPT-5; Hard to say: they perform similarly.} \\

Which evaluation model did you perceive to be more accurate in its judgments of tutor responses – Prometheus or the fine-tuned model from Tab~1 (``Automated Evaluation'')? &
Response options: \textit{The model from Tab~1; Prometheus; Hard to say: they perform similarly.} \\

On a scale from 1 to 5, how easy was it to use the ``LLM Evaluation'' tab? &
Scale: 1 = not easy at all, 5 = very easy.\\
Response options: \textit{1, 2, 3, 4, 5.} \\

Any other feedback is welcome. &
Free-form response. \\

\bottomrule
\end{longtable}

\subsubsection*{Visualizer}

\begin{longtable}{@{}p{0.40\textwidth}p{0.55\textwidth}@{}}
\toprule
\textbf{Item} & \textbf{Question / Response Options} \\
\midrule
\endhead

Instructions &
This tab visualizes statistics from the full dataset of tutorial dialogues and annotated tutor responses. The dataset contains 300 dialogues with responses from 9 tutors (except for the Novice tutor, who has annotations for 76 dialogues). Participants are asked to explore visualizations for at least 2 tutors. \\

On a scale from 1 to 5, how informative did you find these visualizations? &
Scale: 1 = not informative at all, 5 = very informative.\\
Response options: \textit{1, 2, 3, 4, 5.} \\

On a scale from 1 to 5, how easy was it to use the ``Visualizer'' tab? &
Scale: 1 = not easy at all, 5 = very easy.\\
Response options: \textit{1, 2, 3, 4, 5.} \\

We welcome any further feedback. &
Free-form response. \\

\bottomrule
\end{longtable}

\clearpage

\section{User Interface (UI) Details}
\label{app:ui}

\begin{figure}[!h]
    \centering
    \includegraphics[width=\textwidth]{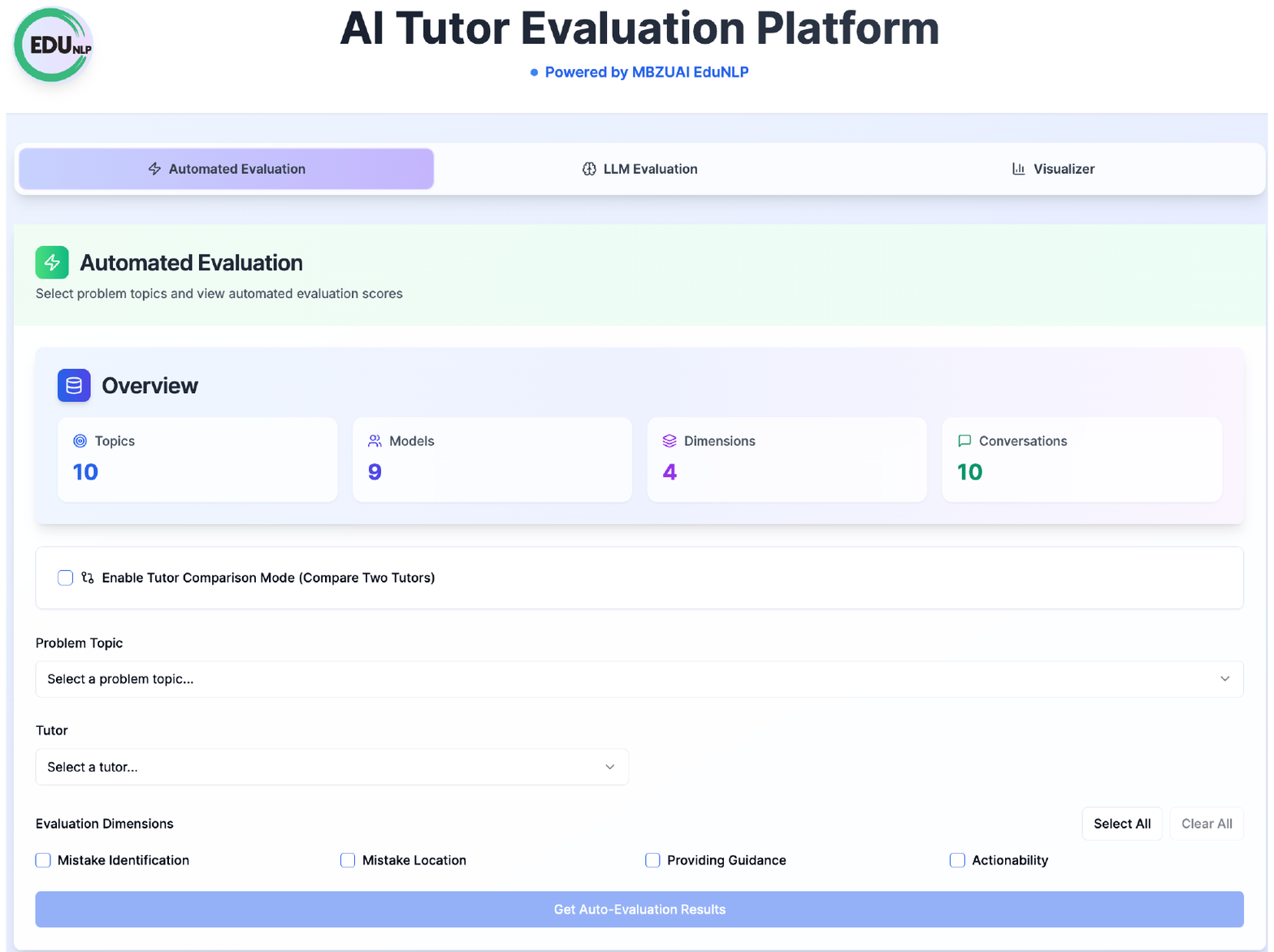}
    \caption{Overview of the UI and the Automated Evaluation Tab.}
    \label{fig:auto_overview}
\end{figure}


\begin{figure}[h]
    \centering
    \includegraphics[width=0.9\linewidth]{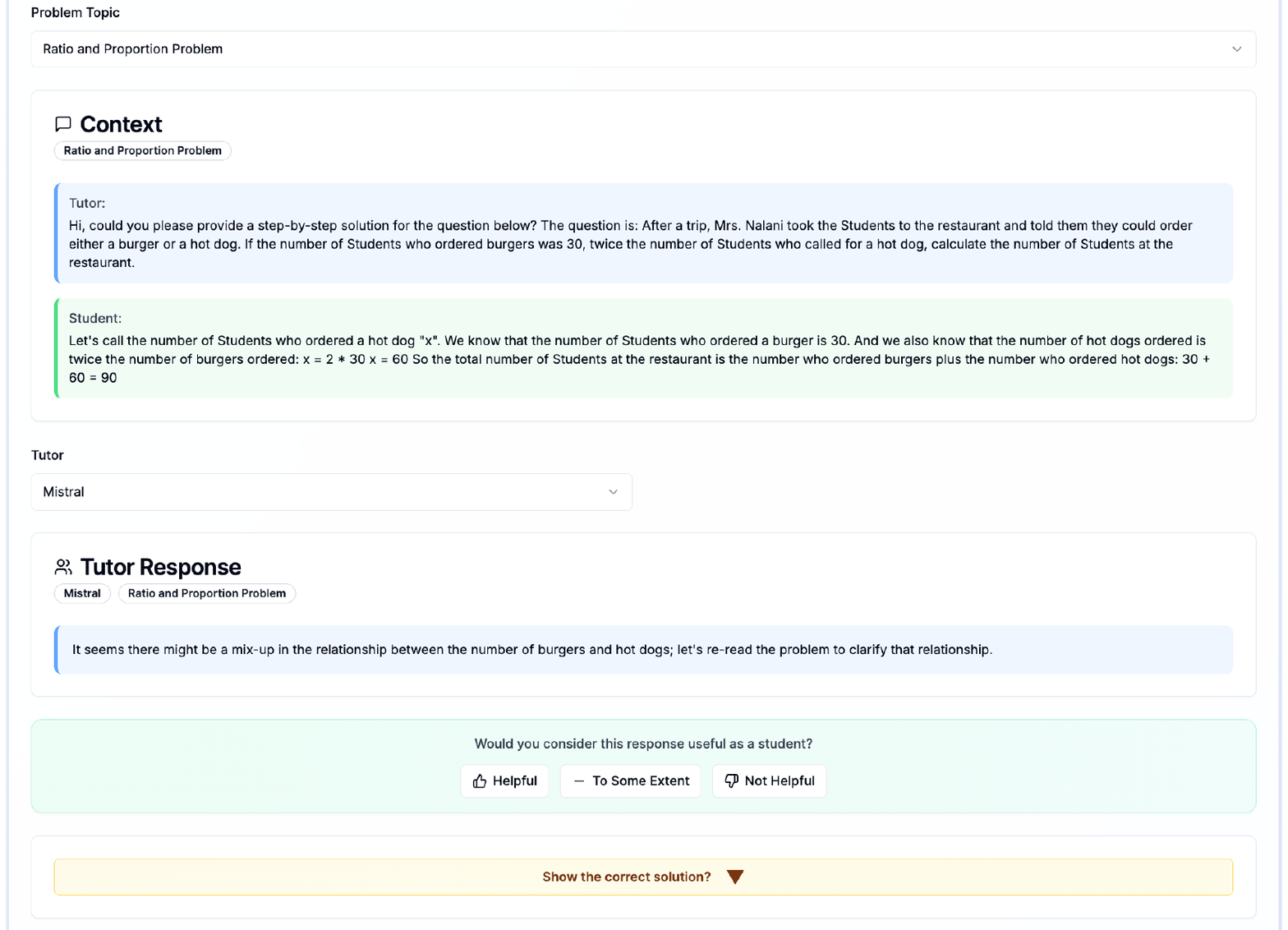}
    \caption{Interface showcasing the selected problem topic with Context, automated Tutor Response, student feedback options, and ground truth verification panel.}
    \label{fig:context_block}
\end{figure}

\begin{figure}[h]
    \centering
    \includegraphics[width=0.9\linewidth]{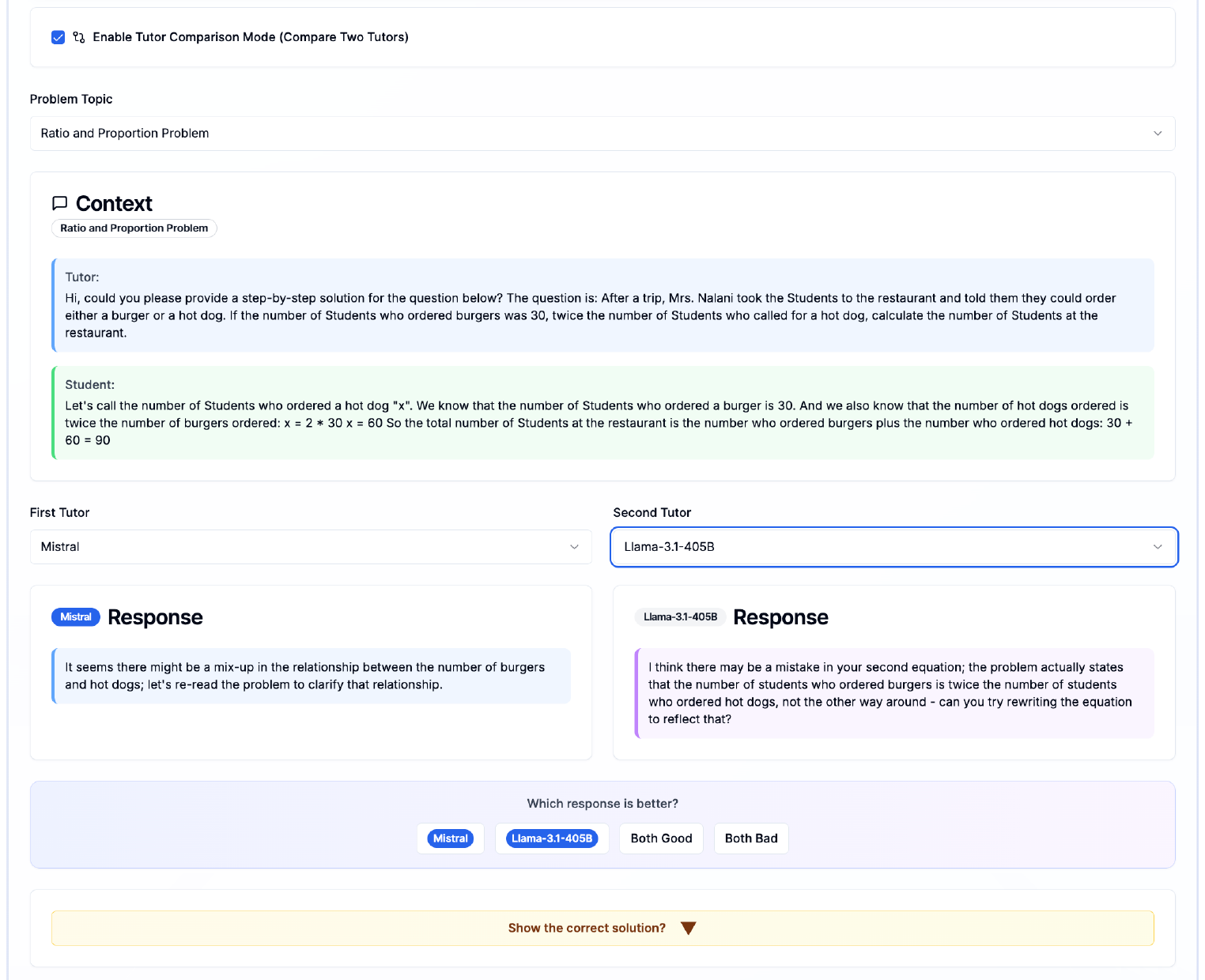}
    \caption{UI displaying the enabled Tutor Comparison Mode, allowing users to compare responses from any two selected tutors for the selected problem topic, along with the Context block, selected tutor responses, feedback options, and ground truth verification option.}
    \label{fig:compare_response}
\end{figure}

\begin{figure}[h]
    \centering
    \includegraphics[width=0.85\linewidth]{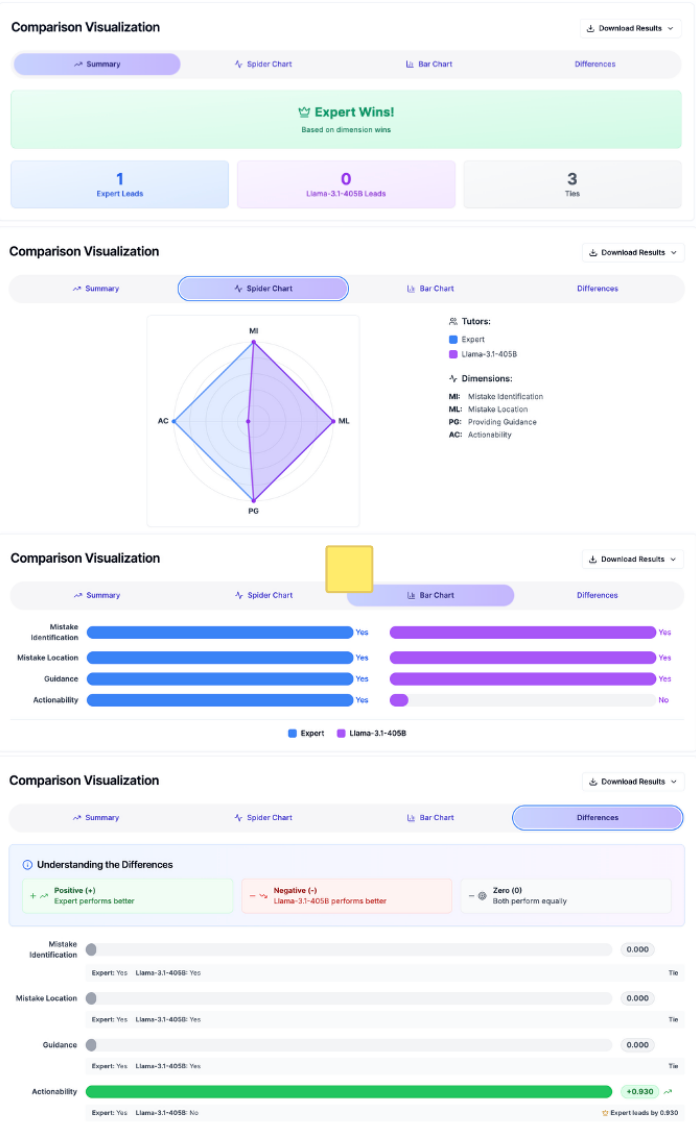}
    \caption{Displayed Tutor Comparison Visualization Panel showcasing Summary metrics, Spider Chart, Bar Chart, and Differences views for the chosen evaluation results. }
    \label{fig:comparison_visualization}
\end{figure}

\begin{figure}[h]
    \centering
    \includegraphics[width=0.8\linewidth]{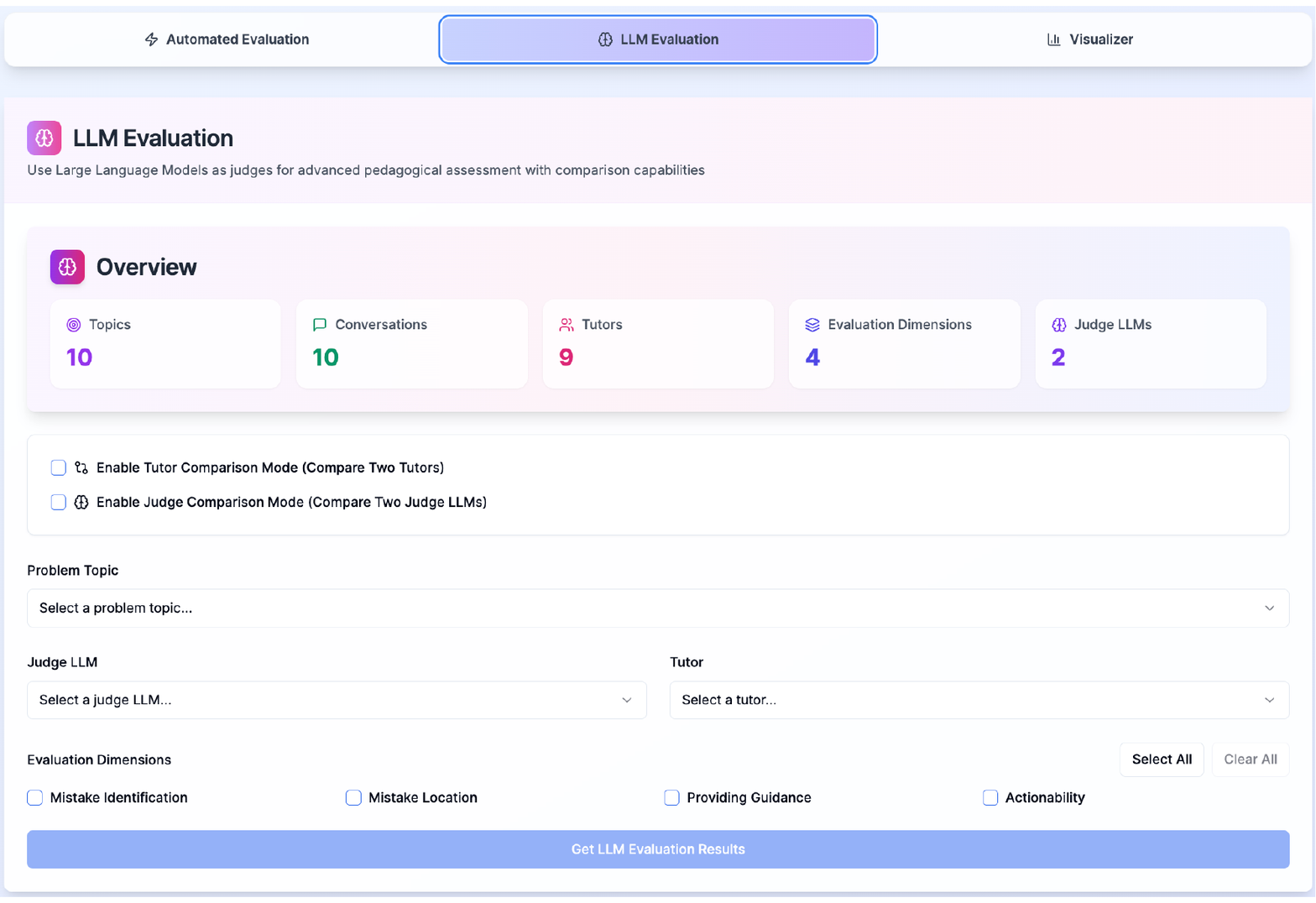}
    \caption{Overview of the LLM Evaluation module showcasing the dashboard panel with statistics on topics, conversations, tutors, and evaluation dimensions. The interface also highlights the provided options for Tutor Comparison Mode and Judge Comparison Mode for advanced pedagogical assessment.}
    \label{fig:llm_eval_overview}
\end{figure}

\begin{figure}[h]
    \centering
    \includegraphics[width=0.9\linewidth]{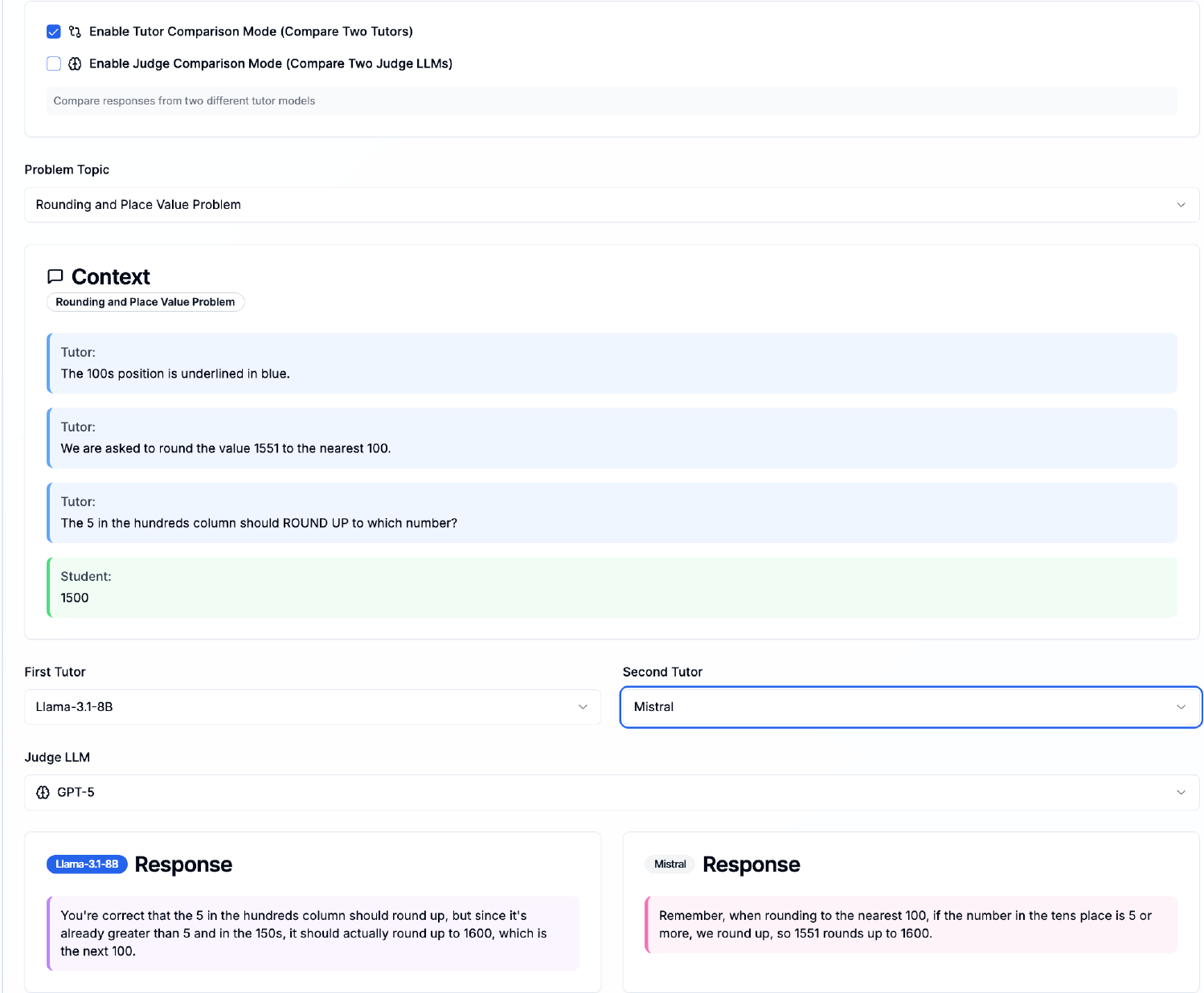}
    \caption{User Interface displaying the enabled Tutor Comparison Mode within the LLM Evaluation module, showing the selected problem, tutor responses from two tutors, and the judge LLM for evaluation.}
    \label{fig:llm_tutor_compare_mode}
\end{figure}

\begin{figure}[h]
    \centering
    \includegraphics[width=\linewidth]{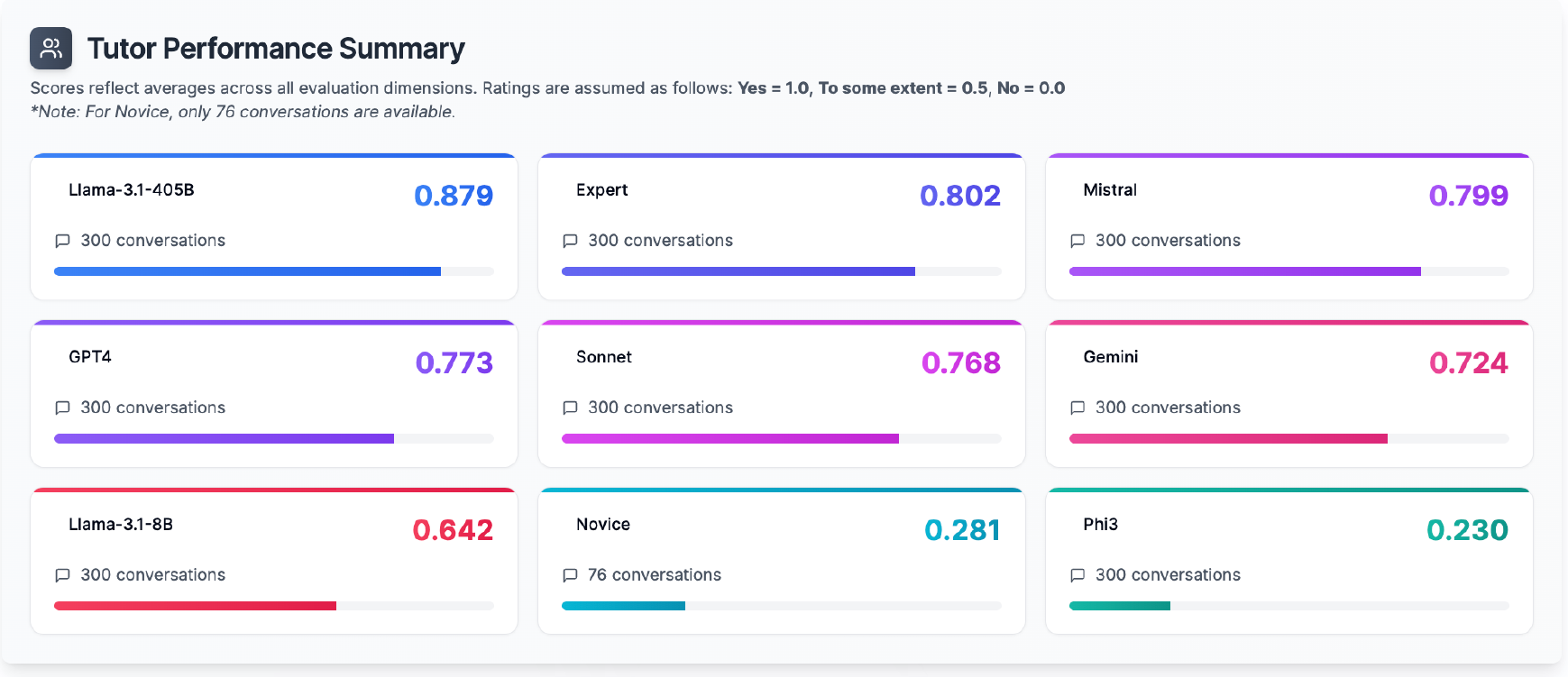}
    \caption{Tutor Performance Summary panel in the Visualizer module, displaying the aggregated evaluation scores for each tutor across all assessment dimensions within the {\tt MRBench} development dataset.}
    \label{fig:tutor_summary}
\end{figure}

\begin{figure}[h]
    \centering
    \includegraphics[width=\linewidth]{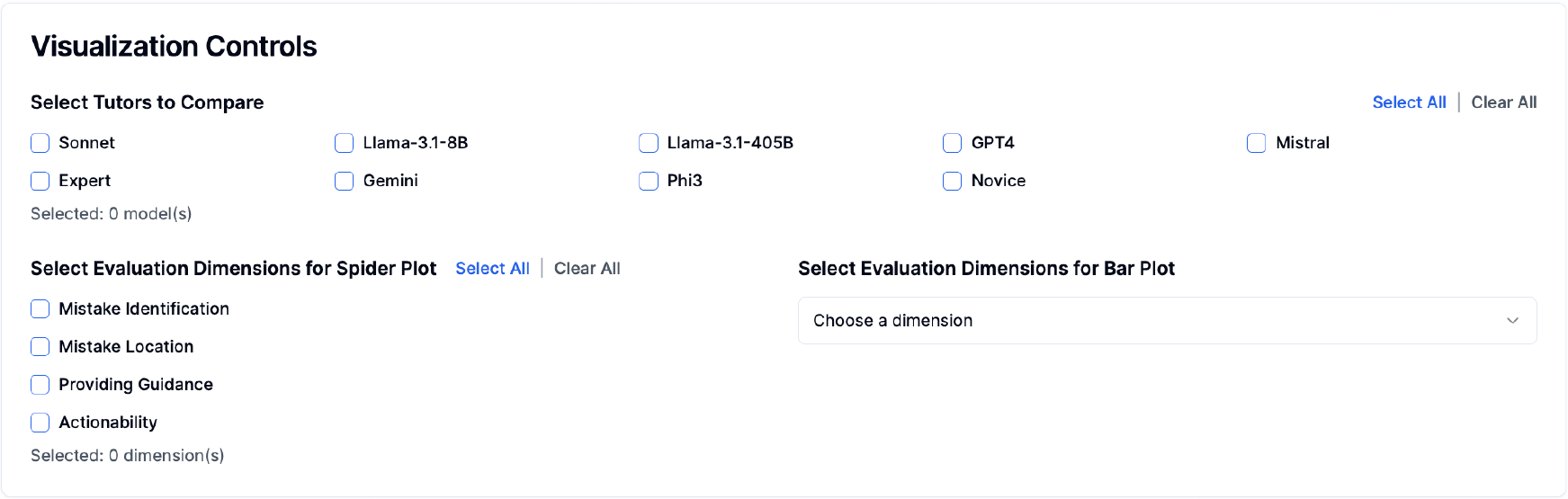}
    \caption{Interface of the Visualization Controls Panel, showing configurable options for selecting tutors and evaluation dimensions to generate comparative spider chart and bar plot visualizations.}
    \label{fig:visualization_control}
\end{figure}

\begin{figure}[h]
    \centering
    \includegraphics[width=0.9\linewidth]{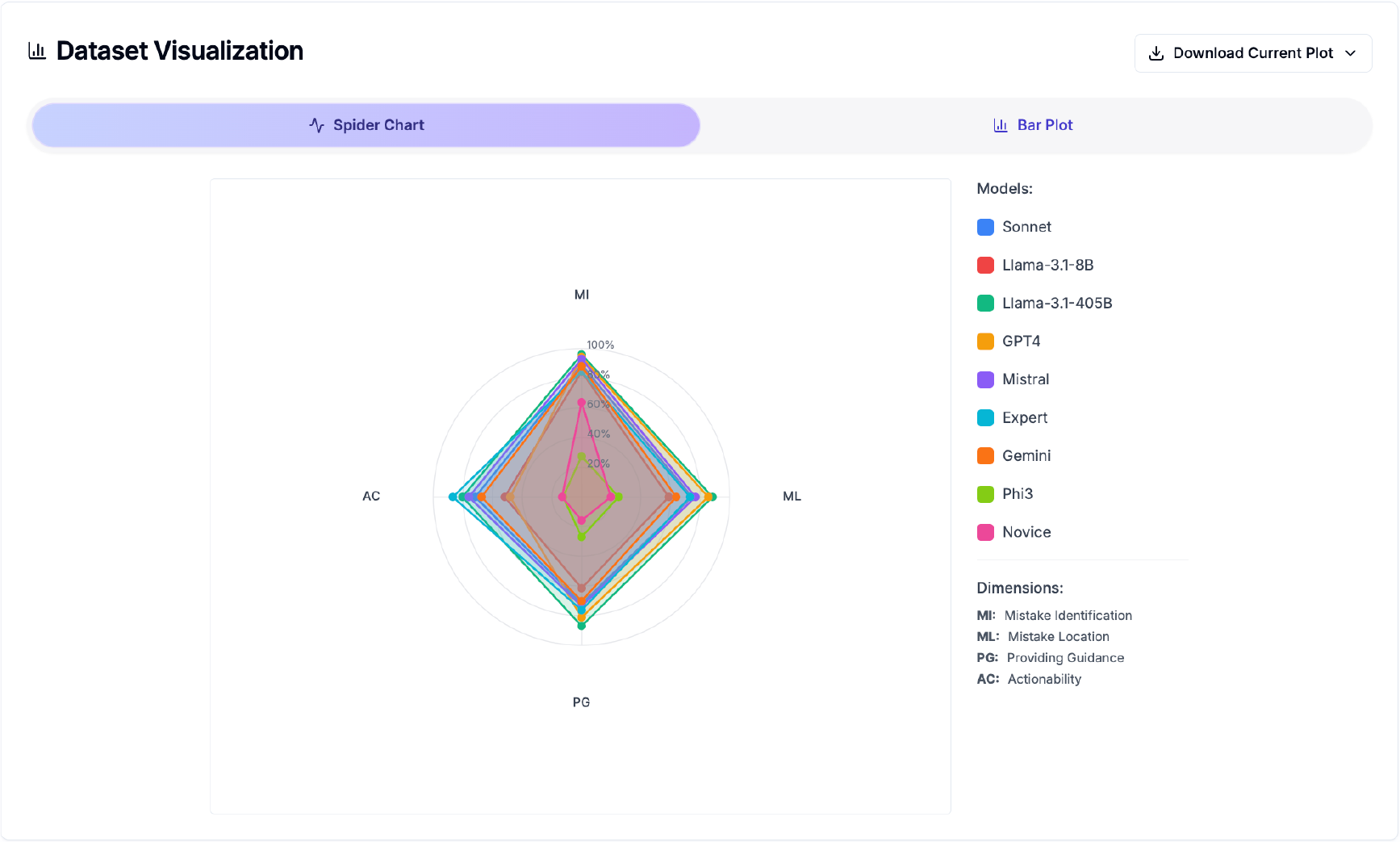}
    \caption{The spider chart representing tutors performance across selected evaluation dimensions, based on configurations chosen in the Visualization Controls Panel.}
    \label{fig:spider}
\end{figure}

\begin{figure}[h]
    \centering
    \includegraphics[width=0.8\linewidth]{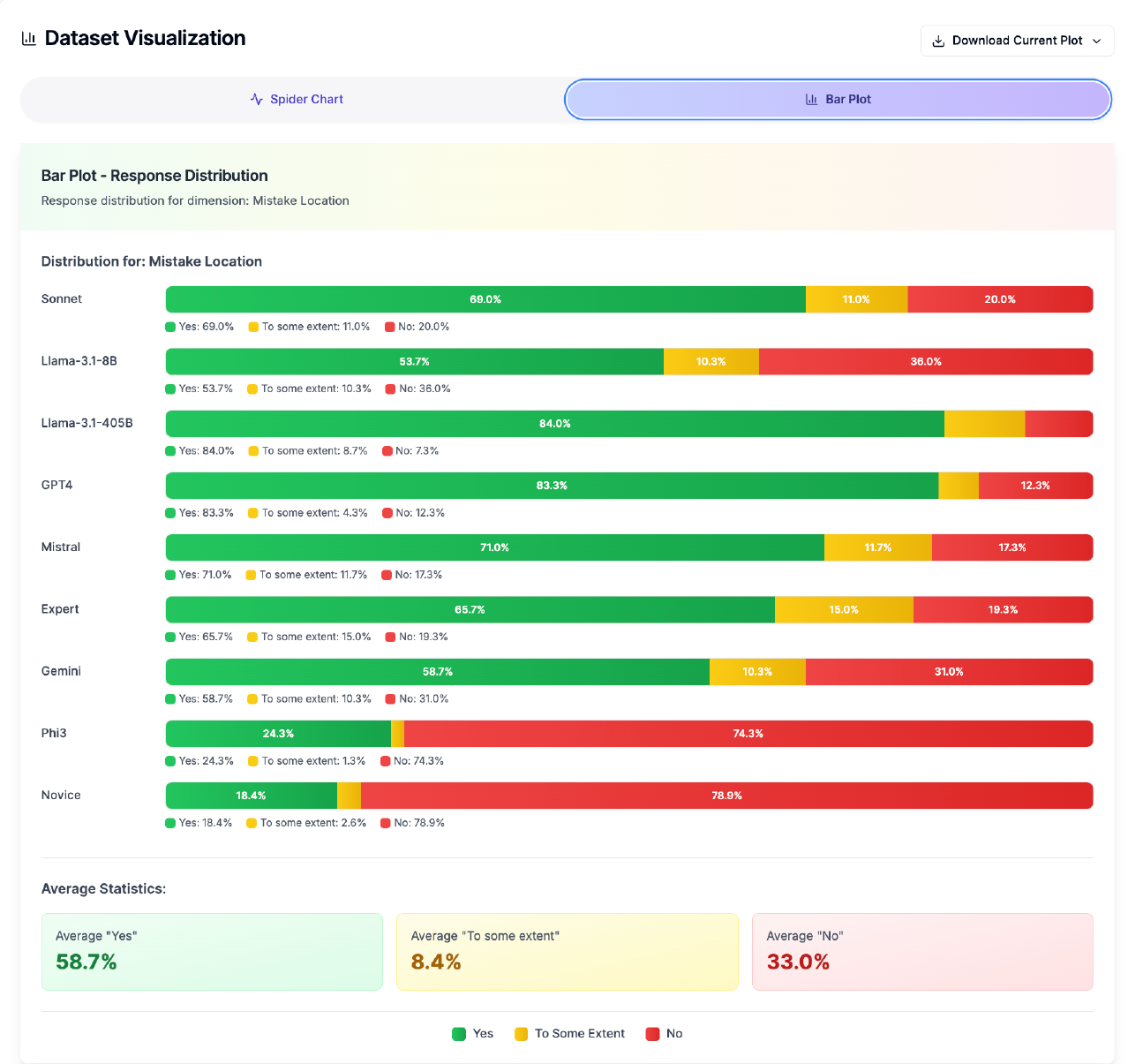}
    \caption{The bar plot representing tutors performance across selected evaluation dimension, based on configurations chosen from the Visualization Controls Panel.}
    \label{fig:bar}
\end{figure}

\twocolumn

\end{document}